\documentclass[10pt,twocolumn,letterpaper]{article}

\usepackage[pagenumbers]{cvpr} 

\usepackage{graphicx}
\usepackage{amsmath}
\usepackage{amssymb}
\usepackage{booktabs}

\usepackage{hyperref}
\hypersetup{pagebackref,breaklinks,colorlinks}

\usepackage[capitalize]{cleveref}
\crefname{section}{Sec.}{Secs.}
\Crefname{section}{Section}{Sections}
\Crefname{table}{Table}{Tables}
\crefname{table}{Tab.}{Tabs.}

\def\cvprPaperID{3406} 
\def\confName{CVPR}
\def\confYear{2023}

\newcommand*{\affaddr}[1]{#1} 
\newcommand*{\affmark}[1][*]{\textsuperscript{#1}}

\begin{document}
\bibliographystyle{unsrt}
\title{Towards an objective characterization of an individual’s facial movements using Self-Supervised Person-Specific-Models}

\author{%
\textbf{Yanis Tazi\affmark[1,2], Michael Berger\affmark[2], and Winrich A. Freiwald\affmark[2,3,4]}\\
\affaddr{\affmark[1]Tri-Institution PhD Program: Cornell, Memorial Sloan Kettering Cancer Center, Rockefeller, NY, NY, USA}\\
\affaddr{\affmark[2]Laboratory of Neural Systems, The Rockefeller University, NY, NY, USA}\\
\affaddr{\affmark[3]The Center for Brains, Minds and Machines, Massachusetts Institute of Technology, Cambridge, MA, USA }\\
\affaddr{\affmark[4]The Price Family Center for the Social Brain, The Rockefeller University, NY, NY, USA}\\
Correspondance to: \textit{yat4003@med.cornell.edu}}
\maketitle

\begin{abstract}
Disentangling facial movements from other facial characteristics, particularly from facial identity, remains a challenging task, as facial movements display great variation between individuals. In this paper, we aim to characterize individual-specific facial movements. We present a novel training approach to learn facial movements independently of other facial characteristics, focusing on each individual separately. We propose self-supervised Person-Specific Models (PSMs), in which one model per individual can learn to extract an embedding of the facial movements independently of the person’s identity and other structural facial characteristics from unlabeled facial video. These models are trained using encoder-decoder-like architectures. We provide quantitative and qualitative evidence that a PSM learns a meaningful facial embedding that discovers fine-grained movements otherwise not characterized by a General Model (GM), which is trained across individuals and characterizes general patterns of facial movements. We present quantitative and qualitative evidence that this approach is easily scalable and generalizable for new individuals: facial movements knowledge learned on a person can quickly and effectively be transferred to a new person. Lastly, we propose a novel PSM using curriculum temporal learning to leverage the temporal contiguity between video frames. Our code, analysis details, and all pretrained models are available in \href{https://github.com/yanistazi/PSM_release}{Github} and Supplementary Materials. 
\end{abstract}

\section{Introduction}
\label{sec:intro}
Faces are rich, complex sources of information. The face conveys structural characteristics, including the identity of its owner \cite{Mckone1294-ij}, along with transient characteristics \cite{Darwin2015-bl,Susskind2008-eh,Frith2009-vc,Adolphs2006-pb,Zadra2011-yr} such as facial expressions using complex individual muscle compositions \cite{Brecht2012-bs}. Of many emotional channels, the face is integral in conveying important information about emotional states \cite{Zadra2011-yr}. Understanding emotions is crucial, as emotions are the bedrock of several life experiences–from survival, to social connections, to well-being \cite{Darwin2015-bl,Plutchik2001-fa,Katana2019-wg}. The human visual system is specialized in the processing of faces and can infer many facial characteristics in fractions of a second \cite{Bar2006-bf,Willis2006-lk}.
\begin{figure}
  \centering
   \includegraphics[width=.9\linewidth]{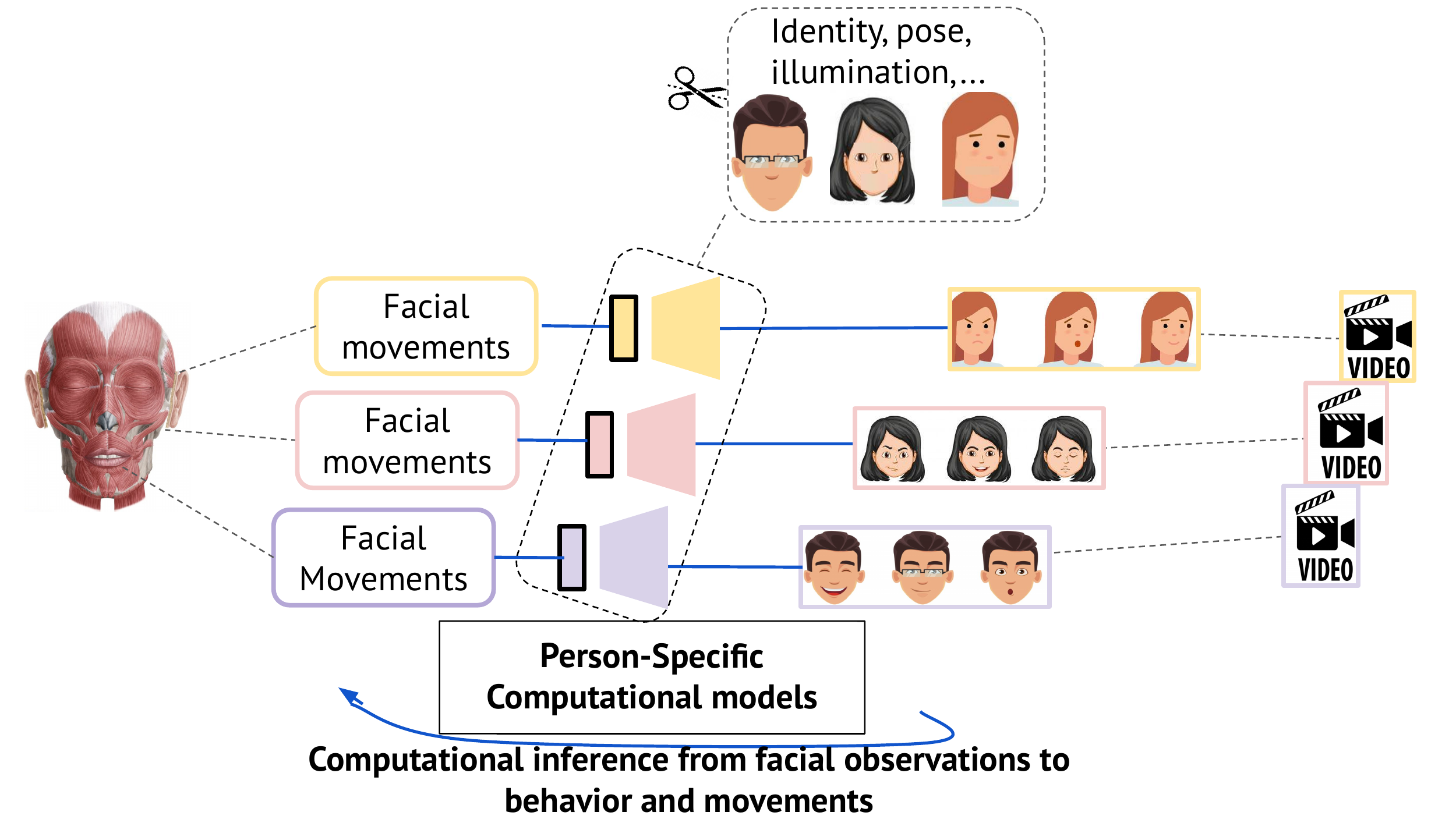}
   \caption{Main idea of the proposed training approach. PSMs learn a person-specific embedding of facial movements from video frames using one model per person.}
   \label{fig:one}
   \vspace*{-2mm}
\end{figure}
Researchers have studied facial movements as a read-out of emotional expression. Several methods have been developed to measure facial movements, such as facial electromyography (fEMG)\cite{Hess2009-xj,Tassinary1992-ve} and Facial Action Coding Systems (FACS)\cite{Ekman_undated-ev}. FACS describes visually distinguishable facial movements using individual components of muscle movement, called Action Units (AUs) \cite{Ekman_undated-ev} (\textit{\cref{tab:tab}}). Unlike EMG tracking, FACS is based on observations from the face annotated by trained human raters.\\
Additionally, data analysis techniques have been developed to recognize facial expressions, such as categorical facial emotion recognition \cite{Kring2007-tn}, and automated computer vision systems that learn facial movements embeddings \cite{Vemulapalli2019-rv,Schroff2015-in}. A meaningful embedding is crucial, as it can be compared with embeddings from other faces to perform tasks like facial expression, facial identity, or facial movement characterization.\\
\begin {table}
\centering
\resizebox{4cm}{!}{
\begin{tabular}{|c|c|}
 \hline
 \textbf{Action Units}& \textbf{Description} \\
 \hline
 1   & Inner brow raiser\\
  \hline
 2 &   Outer brow raiser\\
  \hline
 4 &Brow Lowerer\\
  \hline
 5    &Upper Lid Raiser\\
  \hline
 6&   Cheek Raiser\\
  \hline
 9& Nose Wrinkler\\
  \hline
 12& Lip Corner Puller\\
  \hline
 15& Lip Corner Depressor\\
  \hline
 17& Chin Raiser\\
  \hline
 20& Lip stretcher\\
  \hline
 25& Lips part\\
  \hline
 26& Jaw Drop\\
 \hline 
\end{tabular}
}
\caption {Action Units in the DISFA dataset} \label{tab:tab} 
\vspace*{-3mm}
\end {table}
Computational models of facial behavior emerged in the early 90s and have focused on facial expression recognition (FER) \cite{Mase1991-qc,Cottrell1990-vq}. These automated systems usually share a common framework \cite{Bartlett2011-hy} in which the face is first localized in the image, the information is then extracted from the face region, and this information is leveraged to predict facial expression. These systems have multiple advantages over manual coding , including considerable time gains and scalability to new and large datasets \cite{Sariyanidi2015-mh,Fabian_Benitez-Quiroz2016-hn}.\\
Deep neural networks are now considered standard tools for automated AU detection and FER \cite{Ghayoumi_undated-pi,Mellouk2020-fe}. Several methods have attempted to disentangle face characteristics \cite{Yang2018-ln,Zhang2018-jo}, perform emotion/expression recognition, and execute AUs detection tasks \cite{Jacob2021-oi,Song2021-of,Fan2016-ls,Zakharov2019-ft}. These deep learning methods can be separated into two basic approaches.\\
The first approach is supervised learning, which relies on labeled data consisting of an input and a corresponding label. By training the models to learn to extract the expressions/AUs labels, the model learns to separate the facial expression and its corresponding movements from other facial attributes. These supervised methods rely on large amounts of labeled data, which is expensive, difficult to process in a realistic environment, and are subject to potential errors and biases of initial human labeling.\\
The second approach consists of self-supervised/unsupervised methods \cite{Wiles2018-iw,Shu2018-ov,Xing2019-fk,Chang2021-co}. These methods have attempted to disentangle specific face characteristics using encoder-decoder-like architectures that apply neural networks to reduce the data into low dimensional latent space. The encoder compresses the data into a latent space that represents facial expression and movements, while the decoder converts that latent representation back into a higher dimensional space to reconstruct the input. The decoder ensures that this encoded latent space captures to the greatest extent the facial characteristics of interest. These encoder-decoder-like architectures are particularly useful in learning efficient embeddings from unlabeled data. The intuition is that these low dimensional latent variables are features that should encode the most important information of the facial input to be able to reconstruct it.\\
From a computational perspective, characteristics like pose \cite{Zhang2018-jo,Wiles2018-iw}, shape \cite{Jiang_undated-jy}, or appearance \cite{Shu2018-ov} are separable. In contrast, separating facial expression and its corresponding movements from identity remains very challenging, as individuals exhibit a high degree of variability in facial muscle movements and some of these result directly from differences in facial shape \cite{Kunz2014-jg,Craig2011-xc,Holberg2006-pc}.\\
Very few models have attempted to account for interindividual variability of facial movements. The vast majority of these approaches are supervised and propose to build person-specific classifiers to predict either emotion categories or AU intensities \cite{Sangineto2014-zy,Chu2017-xq,Rescigno2020-ih,Zen2014-zp,Saito2020-vo}. These types of models have been set aside due to a lack of person-specific data and difficulty in model training.\\ 
Even though several computational models of facial expression and characterization of corresponding movements exist, many of these models 1) rely on human labeling assumptions, and 2) do not describe individual differences.\\
In this paper, we propose a novel training approach that can effectively capture an individual’s facial movements independently of its identity and other facial characteristics (\textit{\cref{fig:one}}). We compare state of the art self-supervised approaches in the characterization of facial muscle movements and choose the top-performing architecture to develop self-supervised Person-Specific Models. We demonstrate that these PSMs outperform top-performing approaches in predicting individual facial muscle actions. Most importantly, these models learn an embedding of facial movements that is independent of the person. PSM extracts both facial movements that are shared across persons along with fine-grained behaviors of facial movements characterized by complex patterns of facial actions. These fine-grained facial movements are specific to each individual and not characterized by General Models (GM) trained across individuals that extract more general patterns of facial movements. In addition, we show PSMs are quick to train, and that transfer learning generalizes well across new individuals even across different databases. Lastly, we develop Person-Specific Models with Curriculum Temporal Learning (PSMwCTL) that leverage the temporal order of the frames in a video while accounting for interindividual variability.\\
In summary, our contributions are as follows: 1) self-supervised Person-Specific Models are a novel training approach that focuses on modeling interindividual differences of facial movements,  2) we demonstrate that these models effectively learn a fine-grained representation of facial movements that are specific for each person, allowing for a data-driven characterization of facial movements, 3) the facial movements knowledge that is learned from a person using a PSM can efficiently be transferred to characterize the facial movements of a new person, and 4) we develop a novel temporal curriculum learning approach that increases the complexity of the training examples by leveraging the temporal contiguity of video frames.

\section{Related work}
\subsection{Self-supervised models to learn facial embeddings}
Our objective in developing computational models is to extract an embedding of the facial movements from facial observations that is independent of facial identity and other characteristics without any supervision. While our approach is designed for facial movements in general, we aim to understand facial expressions and we do not test for other movements such as ingestion or speech. The goal is to develop a method that objectively finds such a facial embedding by relying solely on facial observations from video data, while disregarding assumptions underlying human labeling, categorical emotional, and categorical facial expression. 
Self-supervised models are adequate since they can exploit an infinite amount of unlabelled facial video to learn an efficient representation of the facial characteristics. Equally important, these models are fully data-driven and do not rely on human labeling assumptions, and therefore can discover novel facial movements patterns.
We compared 4 recent state-of-the-art self-supervised models that learn facial embedding:\\
1. FAb-Net \cite{Wiles2018-iw} is trained to learn a meaningful face embedding that encodes information about head pose, facial landmarks, and facial expression. The model is optimized to reconstruct a target frame from a source frame by learning the flow field between the two (S.Fig 1A).\\
2. Temporal-consistency \cite{Li_undated-le} leverages the temporal distance of the frames in a video to create a ranking of the frames from an anchor frame. The model is optimized using a temporal ranking triplet-loss to learn a facial embedding (S.Fig 1B).\\
3. Twin-Cycle Autoencoder (TCAE) \cite{Lu_undated-cw} attempts to disentangle all facial movements by separating the head movement motions from the facial muscle movements into two distinct facial embeddings. The model is optimized to change the two main facial movements (head pose movements and expression movements) of the source frame to those of the target frame (S.Fig 1C).\\
4. FaceCycle \cite{Chang2021-co} is trained by overlaying facial motion and facial identity cycle-consistency constraints, separately extracting identity and motion representations when given two facial images of the same person. For the facial motion embedding, the model is optimized to learn the optical flow field from a face with expression to the neutral face of the same person and to reconstruct the original input from the neutral face learned from a different input frame of the same person (S.Fig 1D).\\
These 4 models were trained on images from VoxCeleb1 and VoxCeleb2 datasets \cite{Nagrani2017-xl}. We downloaded the available pretrained models and extracted the trained encoder coding the facial movements as claimed by the authors.
\subsection{Existing models comparison}
\begin{figure}
  \centering
   \includegraphics[width=\linewidth]{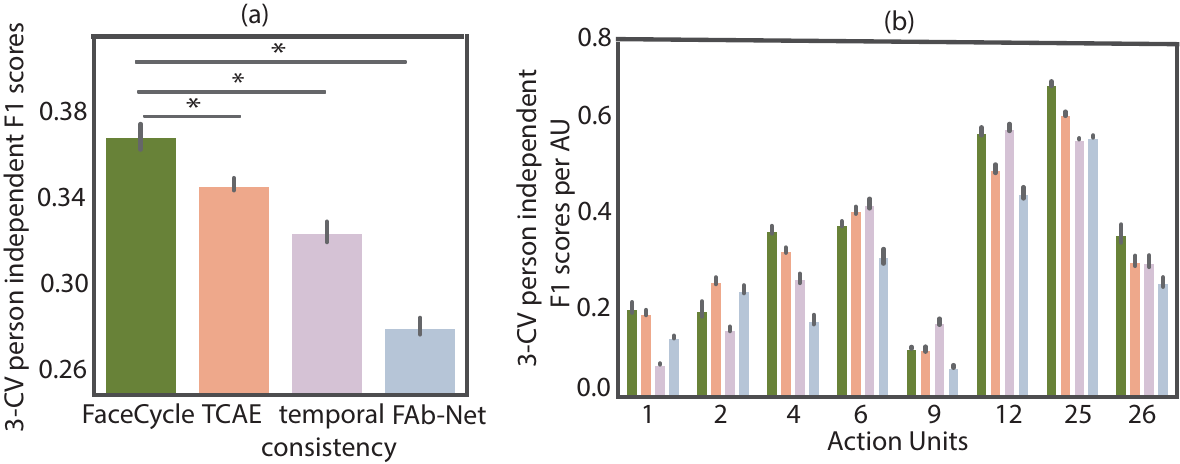}
   \caption {Person-independent F1-score results for state-of-the-art self-supervised methods on DISFA. Higher is better for F1-score. We performed person-independent 3-fold cross validation (3-CV) and we used 100 bootstraps on the test sets. AUs 1, 2, 4, 6, 9, 12, 25, 26 were tested on F1 score results. (a) Results are averaged across AUs. (b) Results are displayed per AU. The pvalues to compare the models were computed with t-test. * denotes $p<0.0001$.}
   \label{fig:two}
   \vspace*{-2mm}
\end{figure}

We adopted the DISFA dataset \cite{Mavadati2013-tc,Mavadati2012} to evaluate the performances of these models. DISFA is particularly adequate to evaluate whether a model learns a good representation of facial movements. It consists of spontaneous video clips of facial behaviors from 27 persons (young adults) across diverse demographics who were video recorded for about 4 minutes. This resulted in $\sim$ 4,800 frames per person watching 9 video segments intended to elicit spontaneous AUs. This dataset provides continuous annotations of spontaneous facial expression coded by experts for 12 facial AUs with intensities ranging from 0 to 5. During the analysis, the frames with intensities greater than 1 were considered as positive, while others were treated as negative. We provide a detailed description of the dataset in S.Fig 2. The analysis of AUs reveals that 1) some action units have higher co-occurrence (for example AU6 with AU12 and 25; AU9 with AU4; AU2 with AU1), which is consistent with facial actions such as smiling and 2) some facial actions are activated more often than others (AU4, 25) (S.Fig 2B-E). More importantly, the analysis highlights the interindividual differences of facial behavior. The individuals have 1) distinct patterns of facial movements, with some individuals more expressive than others (S.Fig 2D), and 2) low temporal correlation of facial movements with some individuals expressing more complex AU patterns than others (average temporal correlation across individuals and AUs = 0.11) (S.Fig 2F). This suggests that even though they are watching the same emotion eliciting videos, their perceived facial actions are different, which again suggests that modeling interindividual differences is essential to uncover the individual’s behaviors of facial movements.\\
We used the standard evaluation protocol \cite{Lu_undated-cw,Li_undated-le} for all 4 models  (FAb-Net, TCAE, temporal-consistency and FaceCycle) by training a linear classifier on the learned embedding. Each model was trained on VoxCeleb and produces a 256 face embedding that contains information for facial actions. Since TCAE produces two embeddings, one for head pose-related features and one for expression-related features, we used the expression-related embedding. Similarly, FaceCycle produces two embeddings and we used the facial-motion embedding. We evaluated the 256 vector representations that were generated by the encoders for AU detection on the DISFA dataset by training a linear classifier (a batch-norm layer followed by a linear fully connected layer with no bias) with Binary Cross Entropy (BCE). We used person-independent 3-fold cross validation (we split the data into 3 folds based on identities) with 100 Bootstraps on the test set using the exact same train/test splits and model parameters (\textit{\cref{fig:two}}). The F1 score (F1 = 2RP/(R+P), was used to report the results as a combination of both precision (P) and recall (R). Our F1 score results are consistent with those previously reported \cite{Lu_undated-cw}. The facial-motion cycle consistency encoder-decoder from FaceCycle by Chang et al. was performing better than the other state of the art models we evaluated. We expected the facial-motion cycle consistency to outperform the other models because the model’s objective is to extract a facial motion embedding from a person’s original input frame. Specifically, the model needs to remove the person’s facial motion to construct its neutral face, and retrieve the person’s facial motion to reconstruct the original face with motion from the neutral face. By doing so, the model learns to implicitly capture the facial motion characteristics, remove them from a given face, and add them back to reconstruct the original input. Although FaceCycle had the best average F1 score amongst evaluated models, FaceCyle’s performances varied for each AU. In particular, for AU2 (outer brow raiser), AU6 (cheek raiser), and AU9 (nose wrinkler), FaceCycle didn’t outperform all other models. This is likely the result of the low proportion and absence of these AUs for some individuals (S.Fig 1 B-C). 

\section{A novel training approach that accounts for individual differences with self-supervised Person-Specific models}
\begin{figure*}[t]
  \centering
  \begin{subfigure}{.45\linewidth}
  \centering
   \includegraphics[width=1\linewidth]{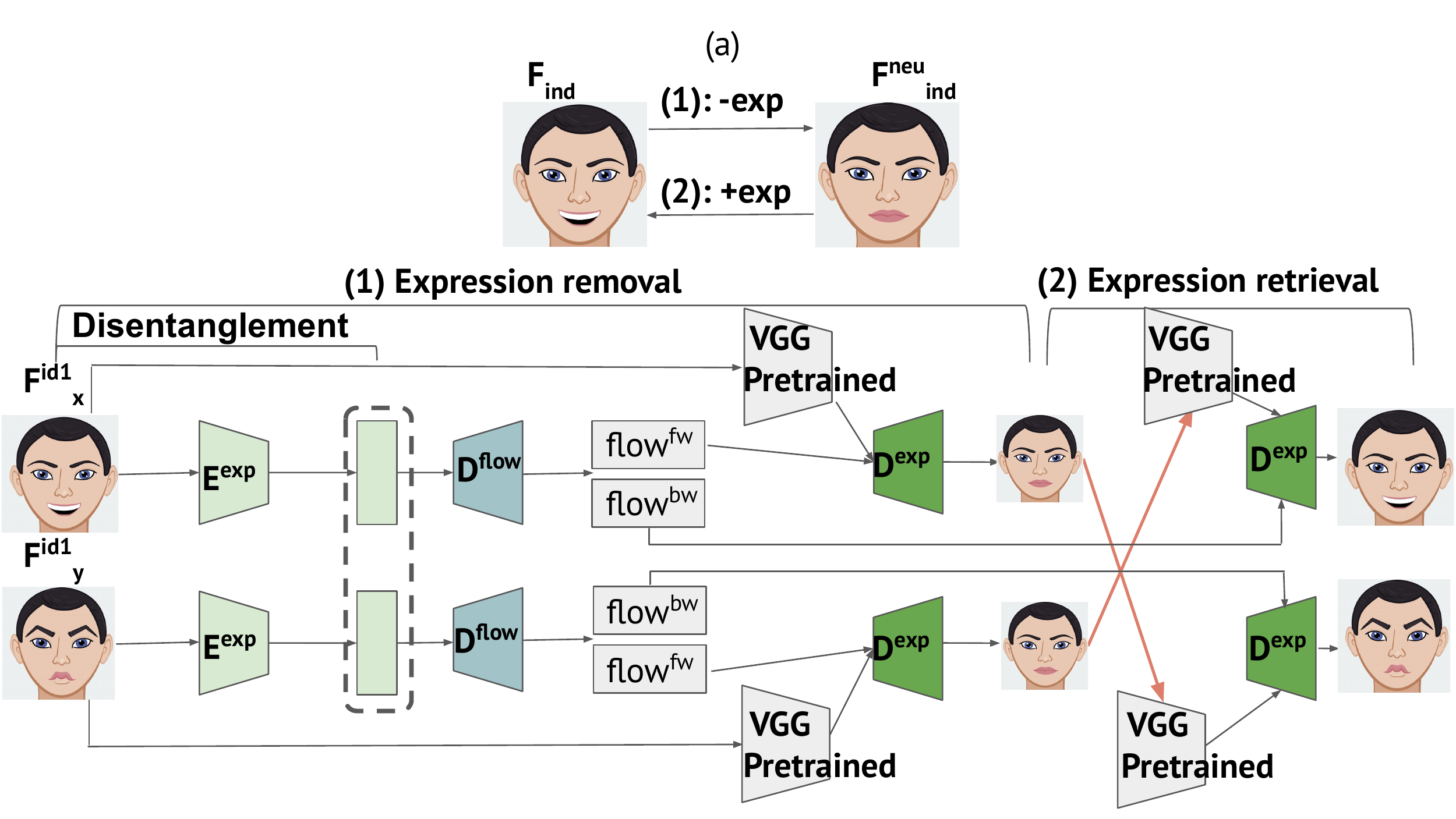}
   \vspace*{-4mm}
  \end{subfigure}
  \begin{subfigure}{.45\linewidth}
  \centering
    \includegraphics[width=\linewidth]{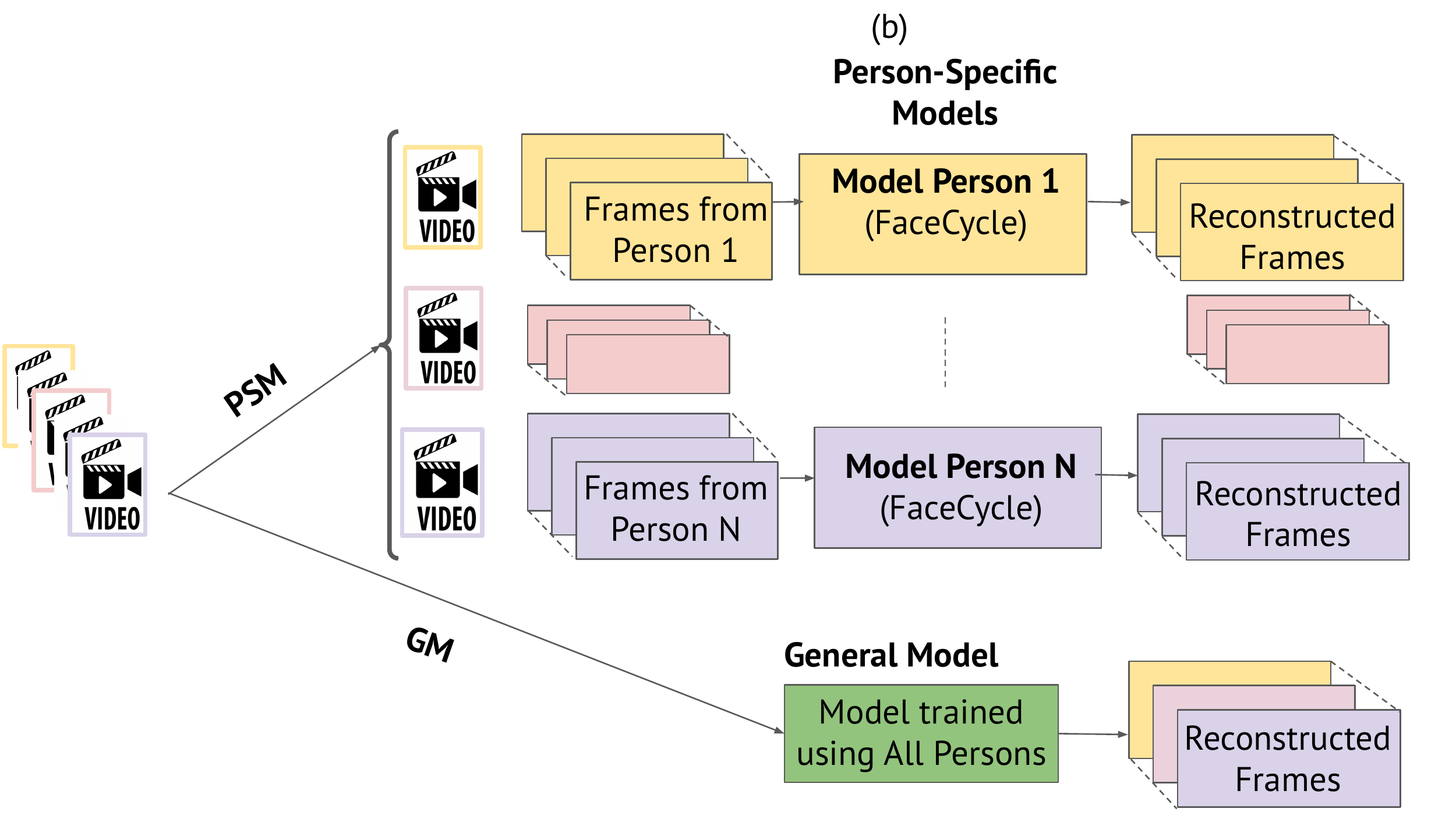}
    \vspace*{-4mm}
  \end{subfigure}
    \begin{subfigure}{.7\linewidth}
    \centering
    \includegraphics[width=.7\linewidth]{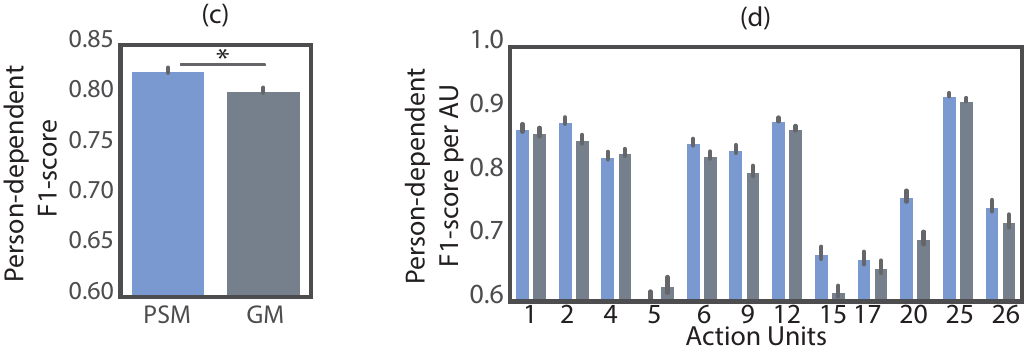}
    \vspace*{-0.5mm}
    \label{fig:3c}
  \end{subfigure}
  \vspace*{-3mm}
  \caption{Comparison between person-specific training approach and standard training approach using the self-supervised Face-Motion Cycle consistency architecture. (a) Main ideas of the facial motion cycle-consistency: de-expression to the neutral face and re-expression to the face with motion. Detailed of the self-supervised architecture with 2 inputs of the same subject. There are 3 main modules: 1.Disentanglement, 2.Expression removal and 3.Expression retrieval.
The dotted rectangle indicates the facial movements embedding representation.
The 2 red arrows are used for the facial-motion cycle consistency invariance constraint. We modified the training optimization process by introducing weight importance decay of the neutral face symmetric loss. For more details about the architecture, please refer to the original paper (Chang et al. 2021)\cite{Chang2021-co} and our code. (b) For PSM, we extract the data for each person and train one model for one individual using the architecture from face-motion cycle consistency. For GM, we trained one model for all persons together. (c) Person-dependent F1-score results for PSMs and GM trained on DISFA. Higher is better for F1-score. Results are averaged across AUs and individuals. (d) Results are displayed for each AUs.\\
Each video was split into 80\% training / 20\% testing while preserving the percentage of samples for each AU class. For each person, AUs that were active more than 2\% of the time were tested and F1 score results were averaged and we used 100 Bootstraps on the test set. The pvalue to compare the 2 training approaches was computed with t-test. * denotes $p<0.0001$.}
\label{fig:three}
\end{figure*}

Even when presented with the same video stimuli, individuals exhibit different facial movements because of interindividual differences (S.Fig 2). Facial movements are highly variables between individuals, which affects the ability of general models to characterize person-specific facial movements. Hence, we introduce a training approach that focuses on Person-Specific modeling. To build person-specific models (PSM) and general models (GM), we use the facial-motion cycle consistency architecture of FaceCycle (\textit{\cref{fig:three}a-b}), the best performing state of the art model evaluated in the previous section. The observed motions are caused by both facial actions and head motions. To get rid of the head movement effects, we first need to localize, crop, and center the face in the image, and then rotate the face such that the eyes lay on a horizontal line. To do that, we used a pre-trained ensemble of regression trees \cite{Kazemi_undated-yl} that estimates 68 facial landmarks positions. We then identify the landmarks corresponding to the eyes and the mouth to crop and rotate the face such that the eyes lay along the same horizontal line. This processing step allows the model to automatically focus on the facial movements rather than head motion. 0.7\% of the frames were discarded because the face was obstructed and the algorithm could not detect the facial landmarks.
We trained PSM on the DISFA dataset with the facial motion cycle consistency in a self-supervised manner with no labels associated. This time, the model learns an embedding that is person-dependent.
We compared the average of all PSMs with the GM. 
\section{Results}
\subsection{PSM outperforms current existing training approach}
We used the trained facial movements encoder for every model to obtain a facial embedding and evaluated the representations generated for facial action unit detection by training a linear classifier with BCE. However, this time, we performed a person-dependent evaluation. We randomly split each person’s video frame into 80\% training / 20\% testing. For each person, we evaluated the F1 score for each action unit, which activated at least 2\% of the time for the person’s video and we used 100 Bootstraps on the test data. All models were trained using the exact same train/test splits and parameters. The person-dependent F1 scores were much higher for all models when compared to the person-independent F1 scores (compare \textit{\cref{fig:three}c} and \textit{\cref{fig:two}a}) for the same linear classifier because the individual’s facial embedding data is now used for both training and testing. This is also a result of the high interindividual variability of the data. The average temporal correlation between participants is only 0.11 (S.Fig 1F) with some participants being more expressive than others (S.Fig 1D).\\
PSM performs better on average than GM (\textit{\cref{fig:three}c}), with both models trained on the same data. PSM predicts all AUs better than GM ($p<0.0001$) except AU4 and 5 (\textit{\cref{fig:three}d}). These F1 score results demonstrate that PSM’s facial embeddings capture more relevant facial movements information than GM’s facial embedding for decoding AUs. Since the face embedding is a higher dimensional space than AU space (256 vs 12 dimensions), it is possible to have different embeddings leading to good AU decoding, and that additional facial movements information is encoded in these embeddings but not in the AUs. We assume that individuals will exhibit behavioral clusters of facial movements. As suggested by the basic emotion theory \cite{Ekman_undated-ev,Darwin2015-bl}, some clusters should be characterized by a clear set of AUs. Among the remaining clusters, some of them might not necessarily be well characterized in AU space: there could be high AU variability within a cluster, or there could be multiple clusters that have similar AU patterns. Some other clusters might be indicative of interindividual differences: they could characterize complex facial movements behaviors specific to particular individuals. In the next section, we further quantify these facial embeddings and compare the clusters found in PSM and GM.
\subsection{PSM learns fine grained movements not characterized by General Models}
\begin{figure}
  \centering
  \begin{subfigure}{.9\linewidth}
   \includegraphics[width=\linewidth]{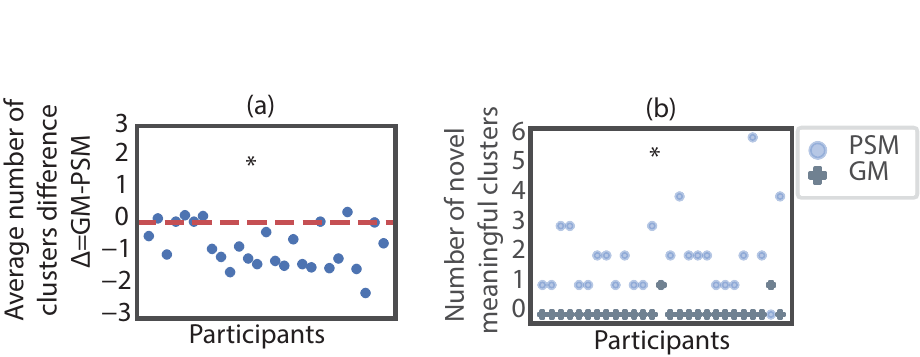}
    \vspace{0.5mm}
  \end{subfigure}
  \begin{subfigure}{\linewidth}
  \centering
    \includegraphics[width=\linewidth]{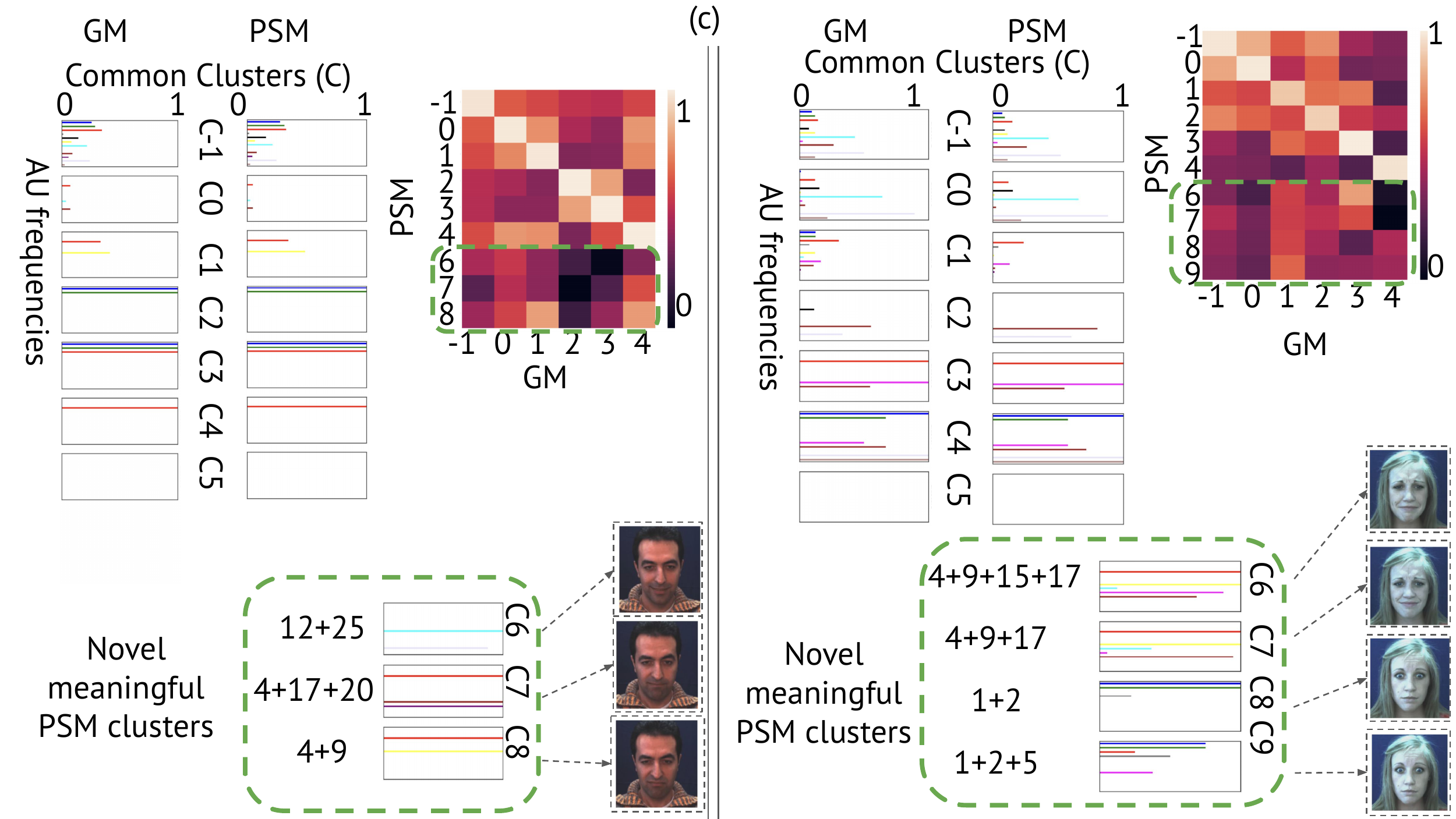}
    \vspace*{-5mm}
  \end{subfigure}
  \caption{Comparison of learned representation of behaviors of facial movements from GM and PSM. \\
(a) Average difference in number of clusters found by density-based spatial clustering of applications with noise clustering (DBSCAN) for each person from PSM and GM. We repeated DBSCAN for different values of $\epsilon$ (3 to 10)) and different values of minimum samples for a cluster to be created (4 to 8) and averaged for each individual. Negative values means that on average, for that individual, the algorithm discovered more clusters from the person-specific model embedding. *:$p<0.0001$\\
(b) Number of novel meaningful clusters for each participant for both PSM and GM. Meaningful cluster is defined using a custom metric leveraging AU frequency correlation ($\rho$) and euclidean distance ($L^1$) between clusters in AU frequency space comparing every cluster in GM and PSM. A meaningful cluster in PSM is a cluster where the custom metric has a value of less than 0.8 with all other clusters in GM, and vice versa. The lower the metric value, the more different the clusters are. Each cluster is defined by a frequency vector of size 12 corresponding to the 12 AU frequencies such that: $PSM_x=(AU1_{prop}^{PSM_x},...,AU26_{prop}^{PSM_x})$, and $GM_y=(AU1_{prop}^{GM_y},...,AU26_{prop}^{GM_y})$:\\ \scalebox{0.85}{$Custom\_metric = Normalized( \rho(PSM_x,GM_y)-L^1(PSM_x,GM_y))$}. *:$p<0.0001$\\
(c) Examples of frames extracted after DBSCAN clustering from the learned embedding by GM and PSM on the frames for 2 participants. Frequencies barplot of AUs are displayed for each cluster. PSM finds more meaningful clusters and these clusters are characterized by specific AU activations. Heatmap corresponds to the normalized value of the custom metric to find meaningful novel clusters. Small values ($<0.8$) consistent across a particular row indicate novel meaningful cluster in PSM. Similarly, small values ($<0.8$) consistent across a particular column indicate novel meaningful cluster in GM. Green rectangle indicates novel PSM clusters.}
    \label{fig:four}
  \vspace*{-3mm}
\end{figure}
In an effort to characterize facial movement clusters and identify the structure of PSM and GM embeddings, we investigated the learned embedding of facial movements using clustering and dimensionality reduction techniques.\\
First, we investigated the number of clusters generated for each person by PSM and GM. We used density-based spatial clustering of applications with noise (DBSCAN) with different parameter values (min samples and $\epsilon$: maximum distance between two samples) and then averaged the number of clusters found for each person for both PSM and GM. We plotted for each person the average difference of number of clusters for GM and PSM. Negative values mean that on average, DBSCAN detects more clusters using the learned embedding from PSM. PSM finds on average one more cluster than GM ($p<0.0001$, values ranging from -5 to +9 clusters difference per person (\textit{\cref{fig:four}a}). However, the results do not indicate whether the clusters found are novel and meaningful in terms of facial movements. Hence, we developed an automatic way to investigate which clusters are novel and well characterized in AU space. We extracted the AU frequencies for each cluster. A PSM cluster is considered novel in comparison to all GM clusters if its AU frequency structure has a low correlation with all other clusters in GM, and a PSM cluster is well characterized in AU space if it contains a high proportion of specific AUs. Hence, we leverage these 2 criterias by combining them into a single metric computing the difference between the correlation and euclidean distance of AU proportions vector for each pair of (PSM,GM) clusters. For each cluster in PSM, we compute this metric for each cluster in GM and vice versa. We then use a threshold of 0.8 and flag clusters that have all pairs below this threshold. This helped us identify novel and well characterized in AU space clusters for each individual in PSM (\textit{\cref{fig:four}b}). We further investigated the content of these clusters for each individual. As an example. we show the results for 2 of these individuals in \textit{\cref{fig:four}c}). GM and PSM both found common clusters with simple behavior characterized by simple AU combinations (for example, for person 1 (left side): cluster with high activation of AU1+2, cluster with high activation of AU4 and cluster with high activation of AU1+2+4). Importantly, we found novel clusters in PSM that were not captured by GM (\textit{\cref{fig:four}c}, S.Fig 3). These clusters are characteristic of fine-grained behaviors of facial movements because they involve high proportions of specific AU combinations and these clusters are not identified by GM, highlighting specific behaviors that are more complex and characterized by combinations of multiple AUs (AU4+17+20, AU4+9). For example, clusters 1, 7, and 8 are characterized by a high activation of AU4, and the differentiation between these 3 clusters involves an additional high proportion of AU17 and AU20 for cluster 7 and AU9 for cluster 8 (\textit{\cref{fig:four}c}). Similar results with different AU patterns hold for all other individuals (S.Fig 3). These behaviors are person specific and could not be characterized by an analysis that looks for general patterns across individuals. On average, PSM finds 2 novel and well characterized behavioral clusters (range 0 to 6) per person that were not identified by GM.\\
Second, the analysis of GM across all individuals revealed facial behavior patterns shared across participants. Clustering analysis of expression embeddings across all individuals revealed that similar expressions and behaviors, characterized by the activation of specific AUs, were grouped together independently of the identities (S.Fig 4).\\
In this section, we demonstrate that the high dimensional embedding learned by these models identify behavioral clusters of facial movements that go beyond action units. These novel clusters identified by PSM are important because they may highlight novel behavioral movements that are specific and unique to one individual or a small group. Previously, these behaviors could not have been characterized by a GM that looks for general patterns of facial expression across individuals. This approach can be of great significance when characterizing behaviors of facial movements of different groups, for instance across different cultures \cite{Elfenbein2002-nm}.
\subsection{PSM transfer learning is effective and can be generalized to new persons}
\begin{figure}
  \centering
  \begin{subfigure}{.45\linewidth}
  \centering
   \includegraphics[width=.95\linewidth]{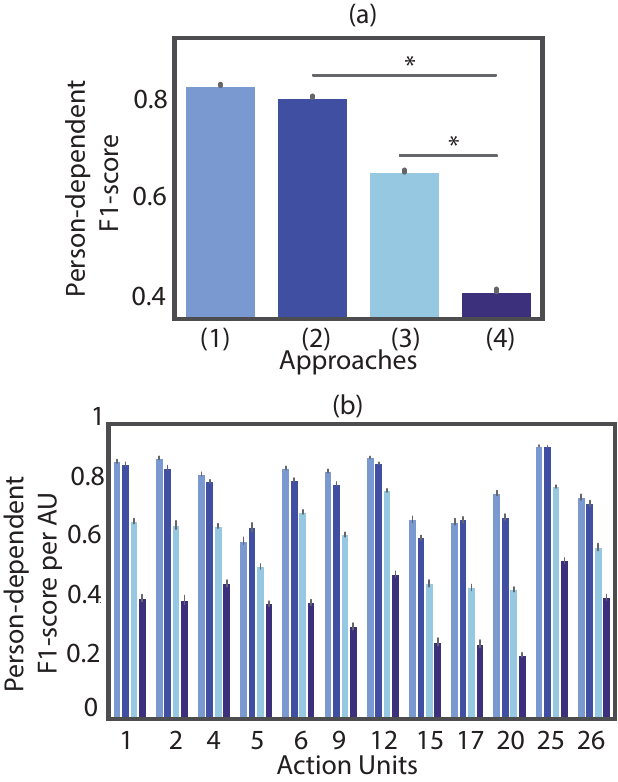}
   \vspace*{-2mm}
  \end{subfigure}
  \begin{subfigure}{.42\linewidth}
  \centering
    \includegraphics[width=.9\linewidth]{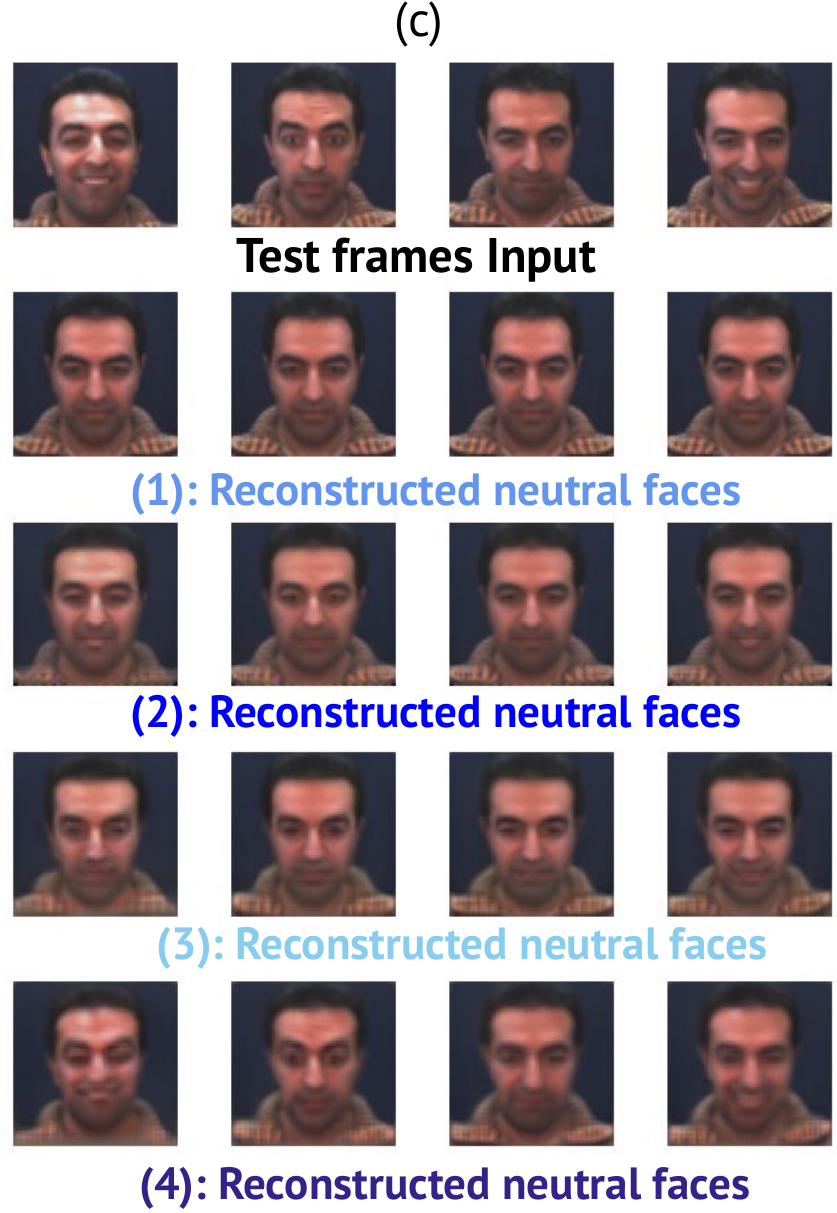}
    \vspace*{-2mm}
  \end{subfigure}
  \caption{Quantitative and qualitative effects of transfer Learning. \\
  (1) We trained non-pretrained PSM for 500 epochs (same results as PSM Figure 3).\\
  (2) We used the VoxCeleb-pretrained GM and trained for 10 epochs. We computed the F1 score results on all test persons and averaged the results.\\
  (3) We trained a non-pretrained PSM on for 500 epochs and then trained it on a new person for 10
epochs.We computed the F1 score results on all possible train/test persons and averaged the results.\\
  (4) We trained a non-pretrained PSM for 10 epochs.\\
  (a) Person-dependent F1-score results for comparing the transfer learning approaches ((2) and (3)) to the upper and lower performance bounds ((1) and (4)). Results are averaged across AUs and individuals. (b) Results are displayed per AUs. The pvalue to compare the 2 training approaches was computed with t-test. * denotes $p<0.0001$.\\
  Each video was split into 80\% training / 20\% testing while preserving the percentage of samples for each AU class. For each person, AUs that were active more than 2\% of the time were tested and F1 score results were averaged and we used 100 Bootstraps on the test set. The pvalues to compare the training approaches were computed with t-test.\\
  (c) Comparison of generated neutral faces for different input frames on the different training approaches.}
  \label{fig:five}
\end{figure}
 \vspace*{-2mm}
Next, we evaluate the generalization of PSMs. PSMs are only practical if they can be trained with limited data and time. We want to determine whether a model trained on one person has learned general features of facial movements that can be directly transferred to encode facial movements for a novel person. To assess the potential of Person-Specific transfer-learning, we used 4 different training approaches: 1) We trained non-pretrained PSM for 500 epochs (same as section 4.1); 2) We used the VoxCeleb-pretrained GM and trained a PSM for 10 epochs; 3) We trained a non-pretrained PSM on for 500 epochs and then trained it on a new person for 10 epochs;  4) We trained a non-pretrained PSM for 10 epochs. We used all frames when training with 500 epochs for approach 1 but only 10\% of the frames when training with 10 epochs. Approaches 1 and 4 provide upper and lower bounds, respectively, for the performances of PSMs trained with limited time. Approaches 2 and 3 tests respectively out-of and within datasets transfer learning. We assessed the transfer learning model both quantitatively and qualitatively.\\
Quantitatively, we used the same training approach as section 4.1 by dividing each person’s video into 80\% training / 20\% testing and trained a linear classifier for each action unit with an activation frequency of at least 2\% and generated 100 Bootstraps on the test data. As before, we average F1 scores across persons. Only for approach 3 for which we apply transfer learning from one person to another, we computed the F1 scores for all possible pairs of train/test persons and averaged the results for each person-dependent F1 scores for each AU. The results demonstrate the effectiveness of transfer learning. Both pretrained models (Approaches 2 and 3) significantly outperformed the model trained for 10 epochs on the person of interest (Approach 4) for each evaluated AUs (\textit{\cref{fig:five}a-b}). We only use 10\% of the frames which corresponds to only 486 frames per individual for pretraining and training for the PSM to illustrate that this approach is scalable and can be applied in real time because the models need very little training and can generalize to new individuals with very little data.\\
Qualitatively, we can assess the effectiveness of transfer learning by looking at the reconstruction of the neutral faces that were held for the test (these tested frames were not in the train dataset). The construction of the neutral face is crucial in the facial motion cycle consistency (\textit{\cref{fig:three}a}), as it is used as the center of the architecture, both as the output for the expression removal and as the input for the expression retrieval. This constructed neutral face is used in the different training stages to train and update the weights of the network. Therefore, the ability to generate a neutral face (i.e to remove the facial-motion from any facial input) is imperative for the network to extract the expression and its corresponding movements independently from a given facial input. Qualitatively, the transfer learning greatly improves the quality of the generated neutral faces (\textit{\cref{fig:five}c}). Transfer learning approaches (approaches 2 and 3) generated more similar neutral faces independently of the initial input when compared to the approach 4 without transfer learning. The average pixel euclidean distance between the generated neutral faces for the transfer learning approaches are respectively 3.0 and 2.8, while the  average pixel euclidean distance is 4.6 for the approach 4 without transfer learning and 0.5 for approach 1 trained for 500 epochs.\\
Lastly, to evaluate the robustness of the models and to ensure that the models are not memorizing neutral faces but actually learning to encode the facial motion and reproduce a given neutral face from any input, we set up a random noise experiment. We generated random noise images and recomputed the generated neutral face. The corresponding generated neutral faces were also random noise (S.Fig 5). This supports that the model does not generate a neutral face from memory.\\
Overall, these results demonstrate that transfer learning is effective and can be generalized to new persons. These results suggest that an individual computational model can quickly learn an independent and objective representation of facial behavioral movements, and that this representation can be transferred to any new individual with very little training. The model can quickly be trained on a person and adapt to new persons with little data (only 10 epochs significantly improve the results using only 10\% of the video frames for that person).
\subsection{Person-Specific Models with curriculum temporal learning (PSMwCTL)}
\begin{figure}[ht]
  \centering
  \begin{subfigure}{\linewidth}
   \includegraphics[width=\linewidth]{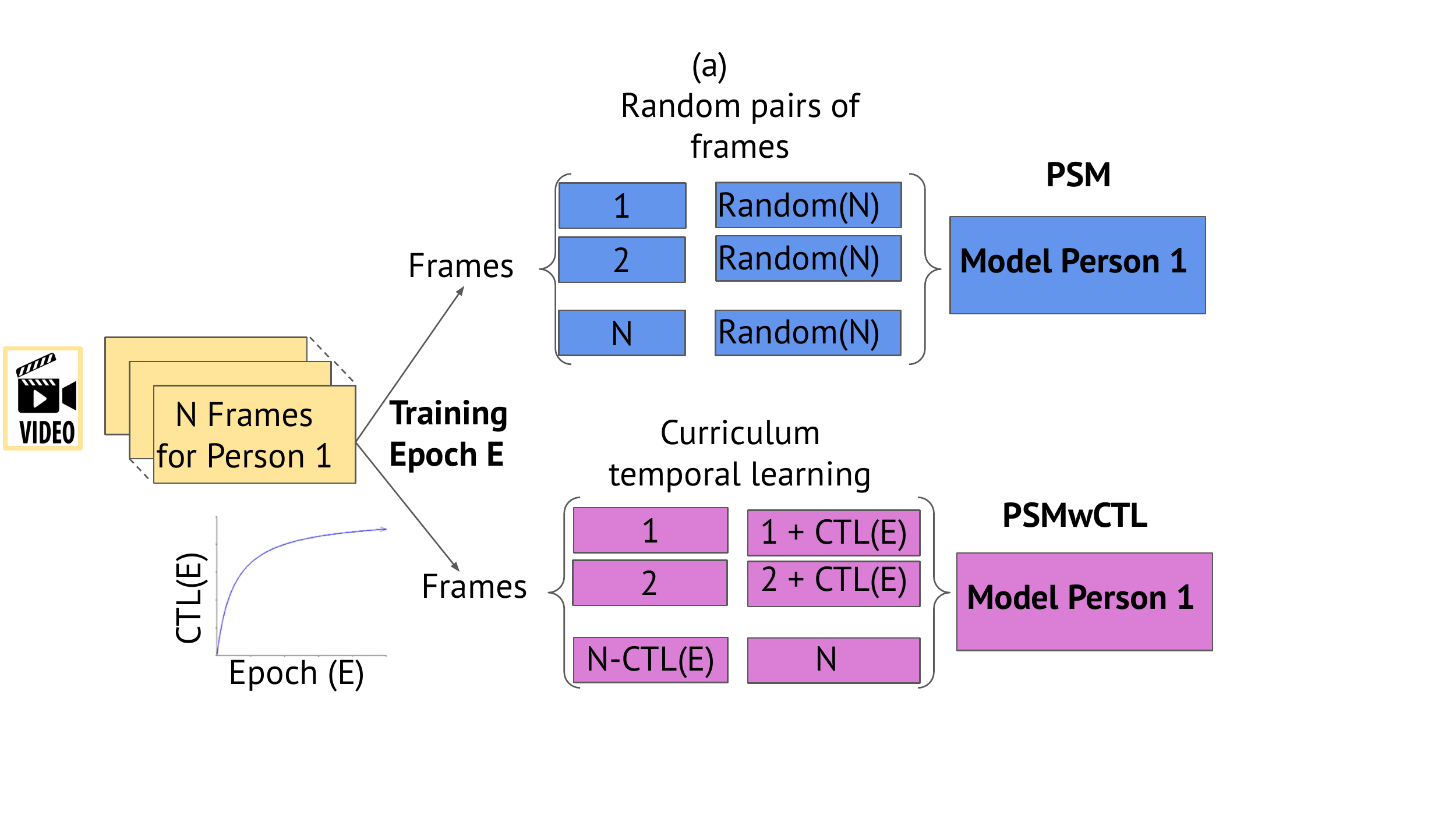}
      \vspace*{-4.5mm}
  \end{subfigure}

  \centering
  \begin{subfigure}{.7\linewidth}
    \includegraphics[width=.7\linewidth]{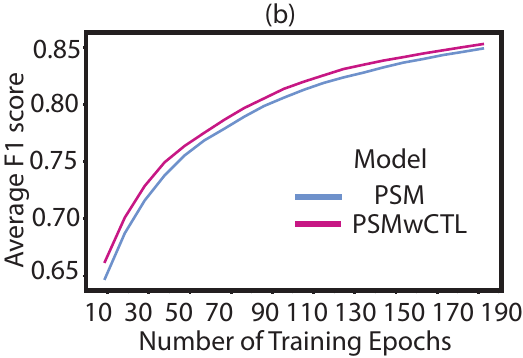}
  \end{subfigure}

  \caption{Introducing PSMwCTL. (a) PSMwCTL framework. See S. Algorithm 1. (b) Average F1 Score for PSM and PSMwCTL over training epochs averaged across AUs and individuals. Each video was split into 80\% training / 20\% testing while preserving the percentage of samples for each AU class. For each person, AUs that were active more than 2\% of the time were tested.}\label{fig:six}  
\end{figure}

Previous results in sections 2 and 3 have demonstrated the potential of person-specific modeling. In addition, temporal-consistency was the second best performing model on person-dependent F1 score. Hence, we developed our modified version of Person-Specific face-motion cycle consistency with curriculum temporal learning (PSMwCTL) by gradually increasing the temporal distance between the two frames used in the face motion cycle consistency for each training example (S. Algorithm 1). This allows us to gradually increase the complexity of the training data by using pairs of frames that get further and further away from each other as the training progresses. Thereby, we leverage the inherent temporal contiguity of the video frames and help our models to learn faster. We evaluate PSM with PSMwCTL and demonstrate that PSMwCTL learns faster than PSM, and the evaluation of AU detection is slightly improved (\textit{\cref{fig:six}a-b}). Once again, we applied the same evaluation protocol by training a linear classifier on the learned embeddings over training epochs for both PSM and PSMwCTL. We plotted the average F1 score over the training epochs. 

\begin{figure}
\captionsetup{labelformat=empty,labelsep=none}
  \centering
   \includegraphics[width=1.05\linewidth]{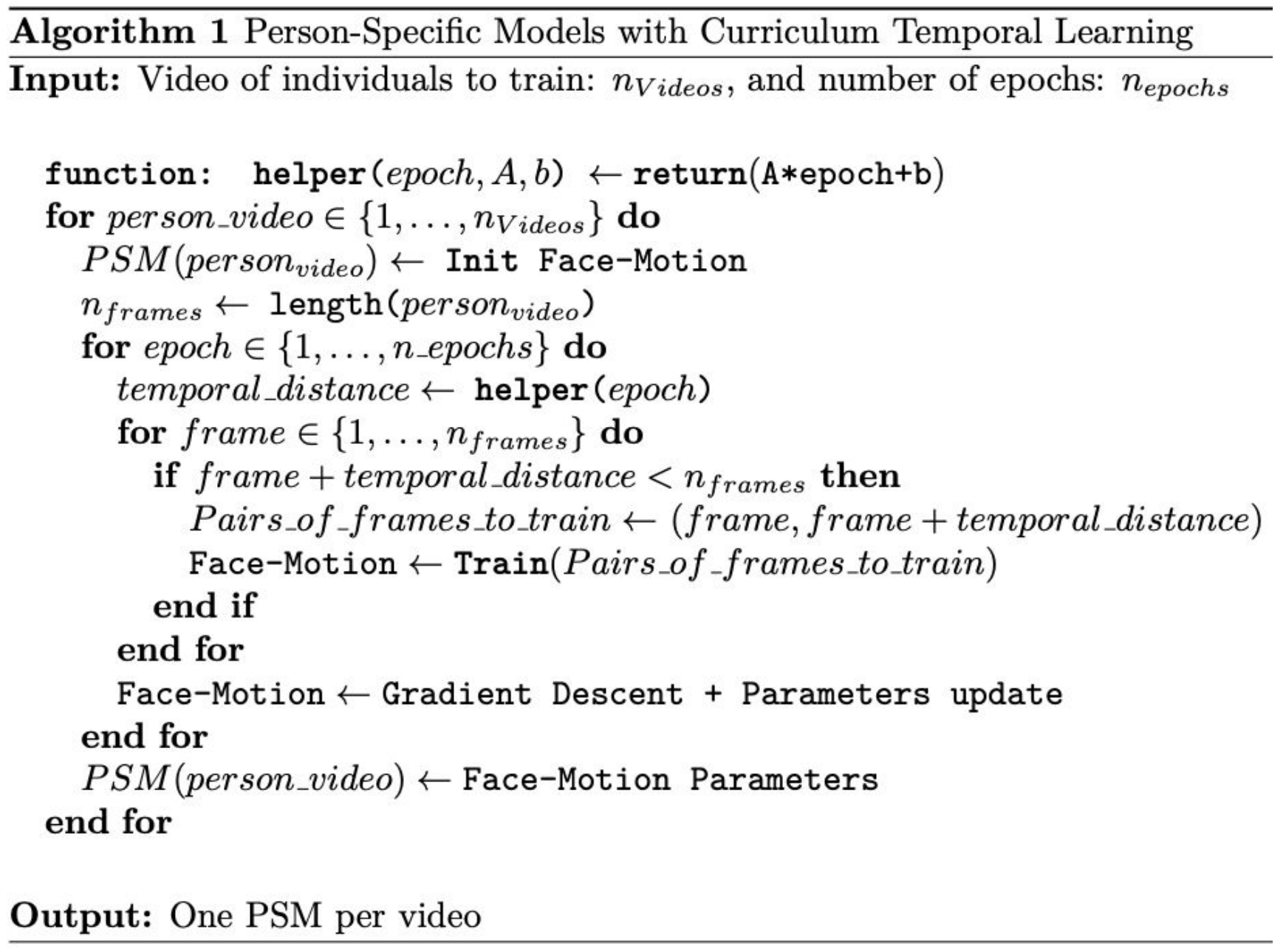}
   \vspace*{-4mm}
\end{figure}

\vspace*{-2mm}
\section{Limitations and future work}
For future work, we plan to build models that can process temporal information in addition to the spatial information to incorporate the dynamics of facial movements and detect subtle facial movements changes. Even though we show that the transfer learning approach generalizes well across different datasets, one main limitation is that the facial dataset we test on is built for one specific context in one specific environment. To make the generalization claim robust, we would need a dataset that encompasses the same participants in different situations, i.e. with a different context and different environment.
\section{Conclusion}
We have proposed a novel training approach for characterizing facial movements that effectively combines facial-motion cycle consistency with Person-Specific self-supervised modeling. By integrating individual differences, our approach can objectively characterize facial muscle movements without relying on human labeling. Our approach demonstrates that the model learns a representation that characterizes a fine-grained behavior specific to each person and highlights that this can not be characterized by large studies with models trained across individuals. With transfer learning, our approach is easily scalable to new persons and can generalize to new datasets. We introduced temporal curriculum learning by leveraging the temporal contiguity of the frames in a video and demonstrated that this technique helps the model learn faster.\\ 
In this paper, we take a step back to objectively characterize facial movements without relying on any human assumptions. By that means, we expect to discover novel behavioral clusters of facial movements. This has several important implications in a world where virtual reality, virtual social interactions, and human-robot interactions are significantly increasing (Metaverse, videoconferencing, etc.). By identifying representations of facial movements that are unique to healthy patients or unique to patients with neurological disorders, this paper opens a new route to identify biomarkers in facial movements for neurological or psychiatric diseases. This positions us to better understand neurological disorders and target neural mechanisms for improved diagnoses \cite{Sheaffer2009-mc,Demenescu2010-pc,Grabowski2019-wf}.\\
\\
\textbf{Acknowledgements}. We thank J.D. Victor, A.J. Hudspeth and C.D. Gilbert for feedback, L.Yin for administrative assistance and the Rockefeller HPC cluster for the GPU support.  
This work was supported by Alexander von Humboldt-foundation (M.B); the Center for Brains, Minds and Machines (CBMM), funded by NSF STC award CCF-1231216 (W.A.F); and by the Department of the Navy, Office of Naval Research under ONR award number N00014-20-1-2292 (W.A.F.). Any opinions, findings, and conclusions or recommendations expressed in this material are those of the author(s) and do not necessarily reflect the views of the National Science Foundation and the Office of Naval Research.\\
\\
\textbf{Contributions}. Y.T., M.B., and W.A.F designed the study, prepared figures and tables, and wrote the manuscript. Y.T with the input of M.B prepared, trained the models and performed all the analysis.

\newpage
{\small
\bibliography{ArXiv_Submission}
}
\newpage
~\newpage
\onecolumn
\section{Supplementary Figures}
\begin{figure*}[h]
\captionsetup{labelformat=empty,labelsep=none}
  \centering
   \includegraphics[width=\linewidth]{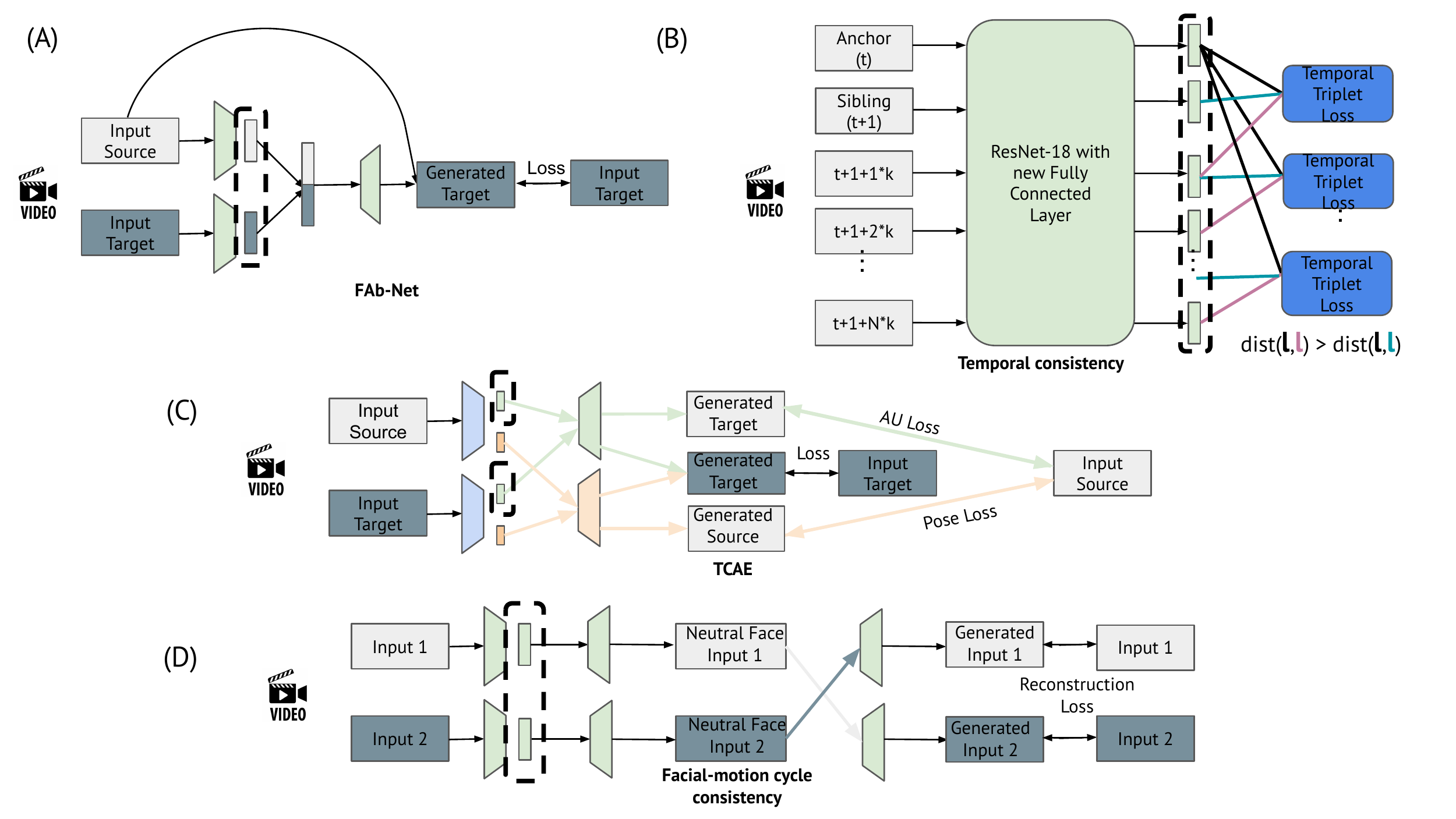}
   \vspace*{-4mm}
   \caption{Supplementary Figure 1. Overview of state-of-the-art self-supervised models of facial embedding.\\
(A) Facial Attributes-Net (FAb-Net), that is trained to embed multiple frames from the same video face into a common low-dimensional space (Wiles et al. \href{https://github.com/oawiles/FAb-Net}{Github}).\\
(B) Temporal-consistency uses a triplet-based ranking approach that learns to rank the frames based on their temporal distance from an anchor frame (Lu et al. \href{https://github.com/intelligent-human-perception-laboratory/temporal-consistency}{Github}).\\
(C) TwinCycle Autoencoder (TCAE) that can disentangle the facial action related movements and the head motion related ones (Li et al. \href{https://github.com/JiaRenChang/FaceCycle}{Github}).\\
(D) Facial-motion cycle consistency from FaceCycle that can disentangle the facial movements by reconstruction of the neutral face (Chang et al. \href{https://github.com/JiaRenChang/FaceCycle}{Github}).Dashed areas indicate the facial-motion embedding.
}
\end{figure*}
\begin{figure*}
\captionsetup{labelformat=empty,labelsep=none}
  \centering
   \includegraphics[width=1\linewidth]{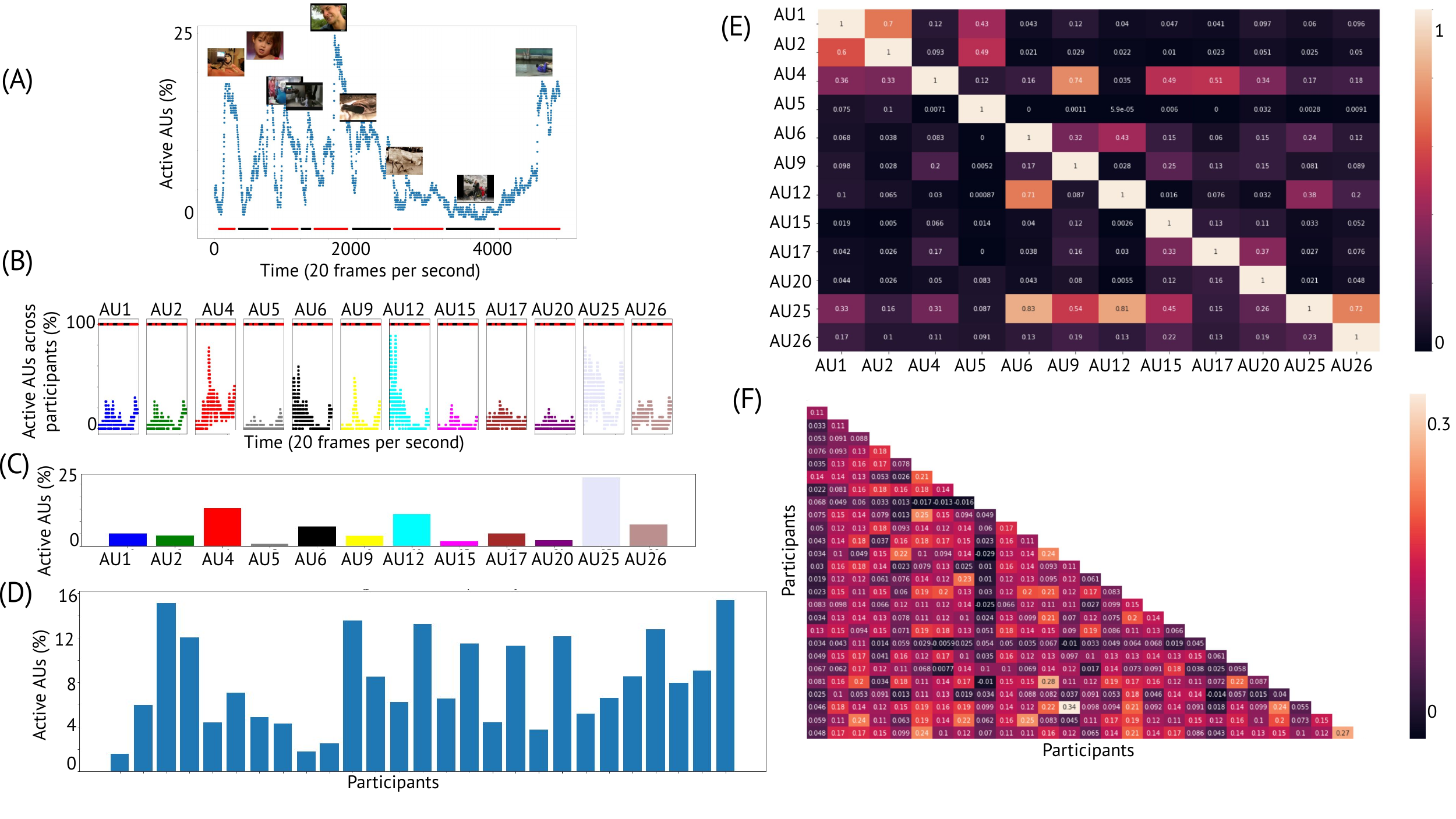}
   \vspace*{-4mm}
   \caption{Supplementary Figure 2. Detailed analysis of the facial muscle movements in the DISFA dataset through facial action units (AUs). 
AU intensities greater than 1 were considered as positive, while others were treated as negative.\\
(A) Scatter plot of average number of active AUs overtime across participants. Average is calculated across AUs and across participants for each frame. x axis represents the duration of the video (20 frames per second) and the y axis represents the average number of active AUs across participants for a particular frame. The images represent selected frames from the video stimuli. The alternance between red and black line corresponds to the different Youtube video segment stimuli.\\
(B )Scatter plot of average frequency for each AU overtime across participants. Each panel represents a different AU.\\
(C) Barplots of average frequency for each action unit across participants and time.\\
(D) Barplots of average frequency of action units across time for each individual.\\
(E) Co-occurrence of action units across participants. Each column represents the likelihood of each AU being activated when the AU corresponding to that column is activated.\\ 
For example, $P(AU2 = 1|AU1 = 1)=0.6$, i.e when AU1 is activated, the likelihood of AU2 being activated is 0.6.\\
(F) Average temporal correlation across action units for the participants.
}
\end{figure*}
\begin{figure*}
\captionsetup{labelformat=empty,labelsep=none}
  \centering
   \includegraphics[width=1\linewidth]{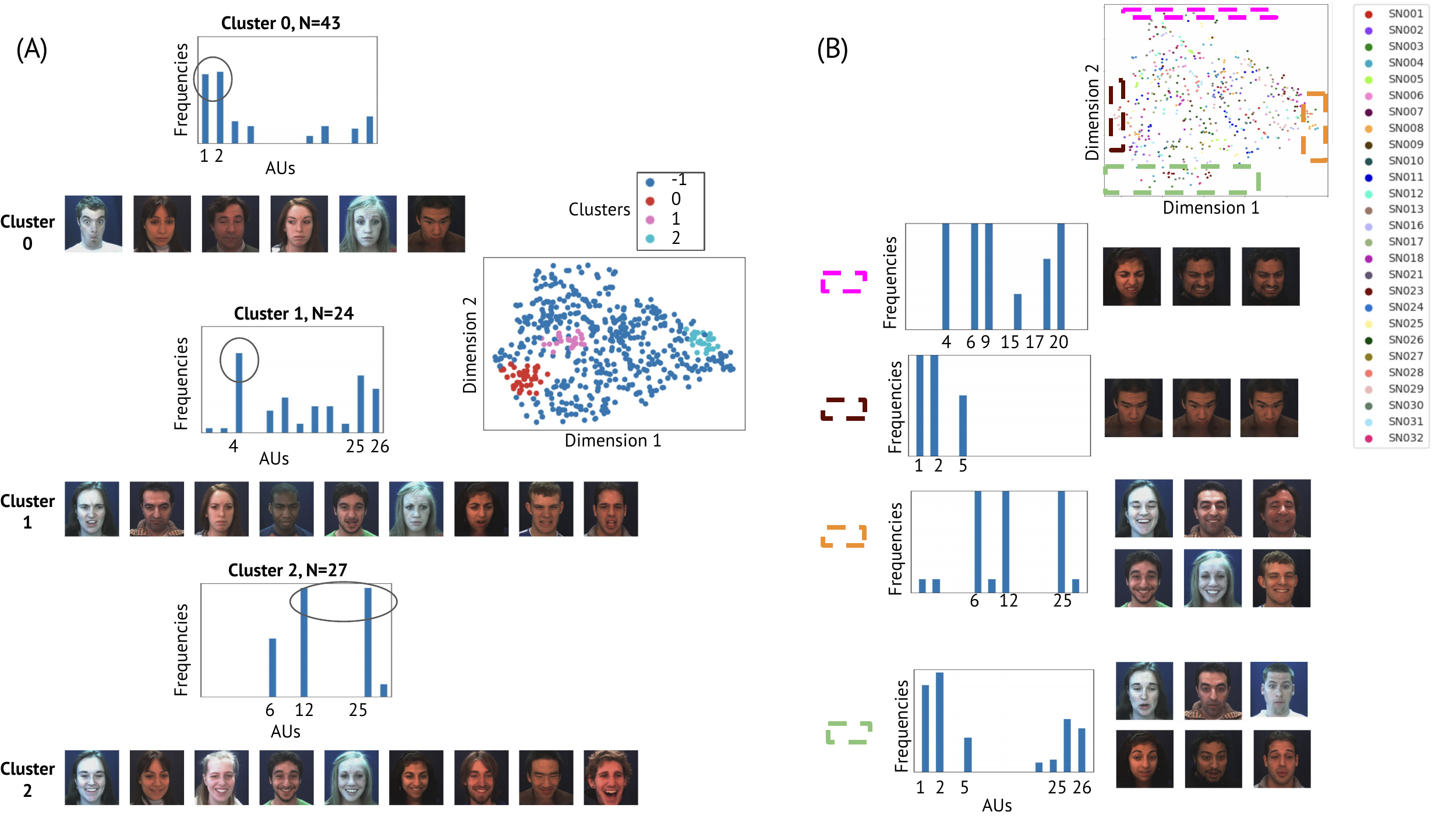}
   \vspace*{-4mm}
   \caption{Supplementary Figure 4. Facial behavior embedding analysis learned from GM.\\
(A) Examples of frames extracted by density-based spatial clustering of applications with noise (DBSCAN) analysis from the GM learned embedding across persons.
Cluster 0 reveals high activation for AU1 and AU2, cluster 1 reveals high activation for AU4, and cluster 2 reveals high activation for AU12 and AU25. Axis 1 and 2 projects on the 2 first principal dimensions.\\
(B) Examples of behaviors of facial movement encoded by the learned embedding from the General Model.\\
Dimensions 1 and 2 projects on the 2 first principal dimensions. Frequencies barplot of AUs are displayed for each cluster and rectangle areas.
}
\end{figure*}
\begin{figure*}
\captionsetup{labelformat=empty,labelsep=none}
  \centering
   \includegraphics[width=1\linewidth]{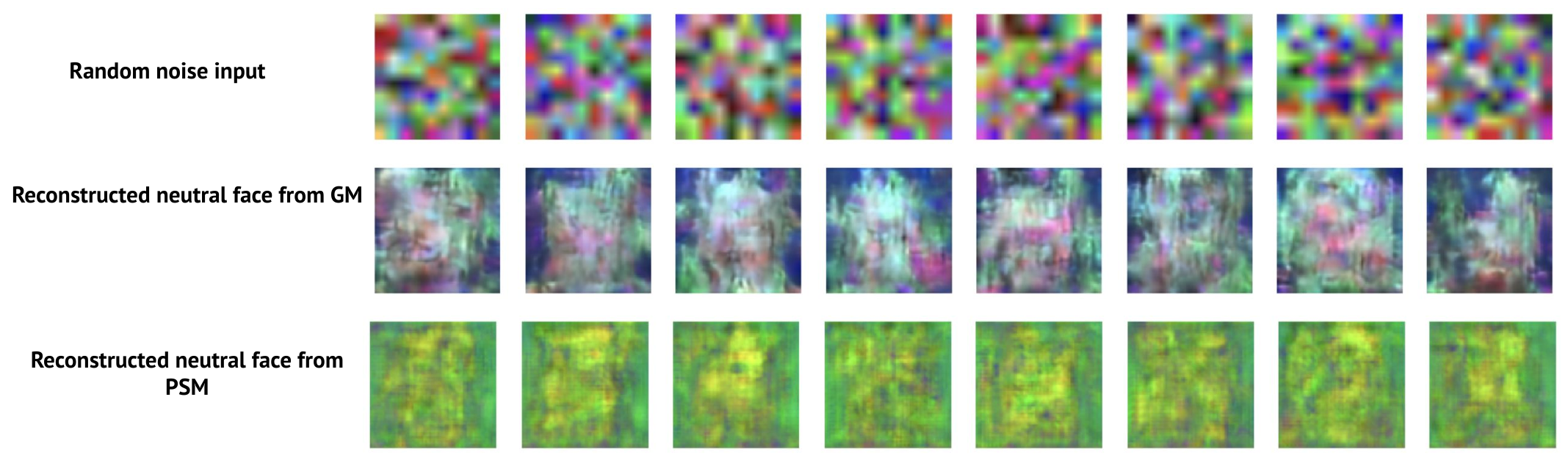}
   \vspace*{-4mm}
   \caption{Supplementary Figure 5. Reconstruction of the neutral face from random noise experiments from both General and Person-Specific Models.\\
Generated neutral faces corresponding to random noise input were also random noise.
}
\end{figure*}
\clearpage
\begin{figure*}
\captionsetup{labelformat=empty,labelsep=none}
   \caption{Supplementary Figure 3. Clustering analysis in AU space from the learned embedding by GM and PSM and discovery of novel meaningful clusters for all remaining participants.\\
Frequencies barplot of AUs are displayed for each cluster. PSM finds more meaningful clusters and these clusters are characterized by specific AU activations. Heatmap corresponds to the normalized value of the custom metric to find meaningful novel clusters. Small values ($<0.8$) consistent across a particular row indicate novel meaningful cluster in PSM. Similarly, small values ($<0.8$) consistent across a particular column indicate novel meaningful cluster in GM. 
}
\end{figure*}

\begin{figure*}
\captionsetup{labelformat=empty,labelsep=none}
  \centering
   \includegraphics[width=1\linewidth]{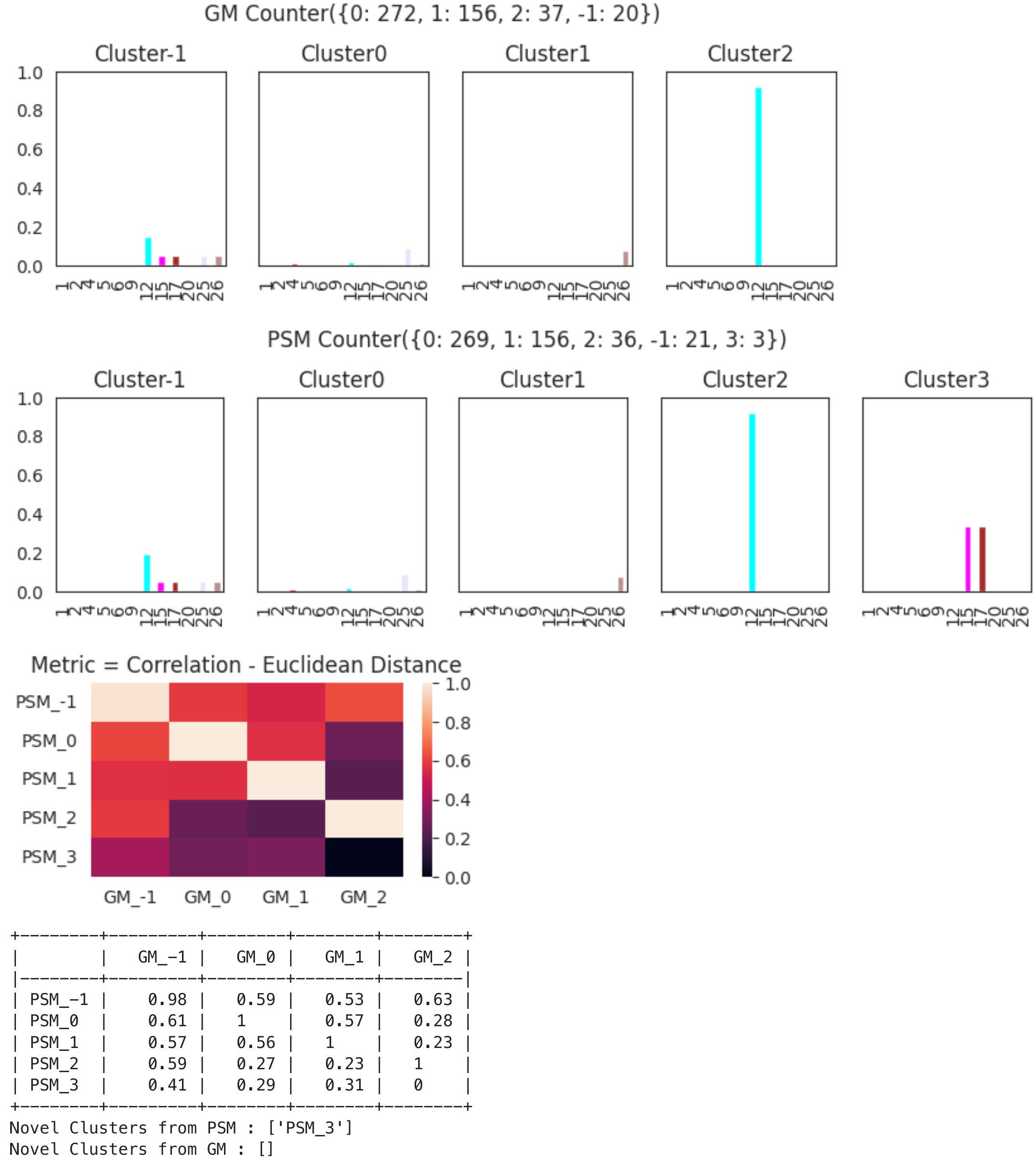}
   \vspace*{-4mm}
   \caption{SN001}
\end{figure*}
\begin{figure*}
\captionsetup{labelformat=empty,labelsep=none}
  \centering
   \includegraphics[width=1\linewidth]{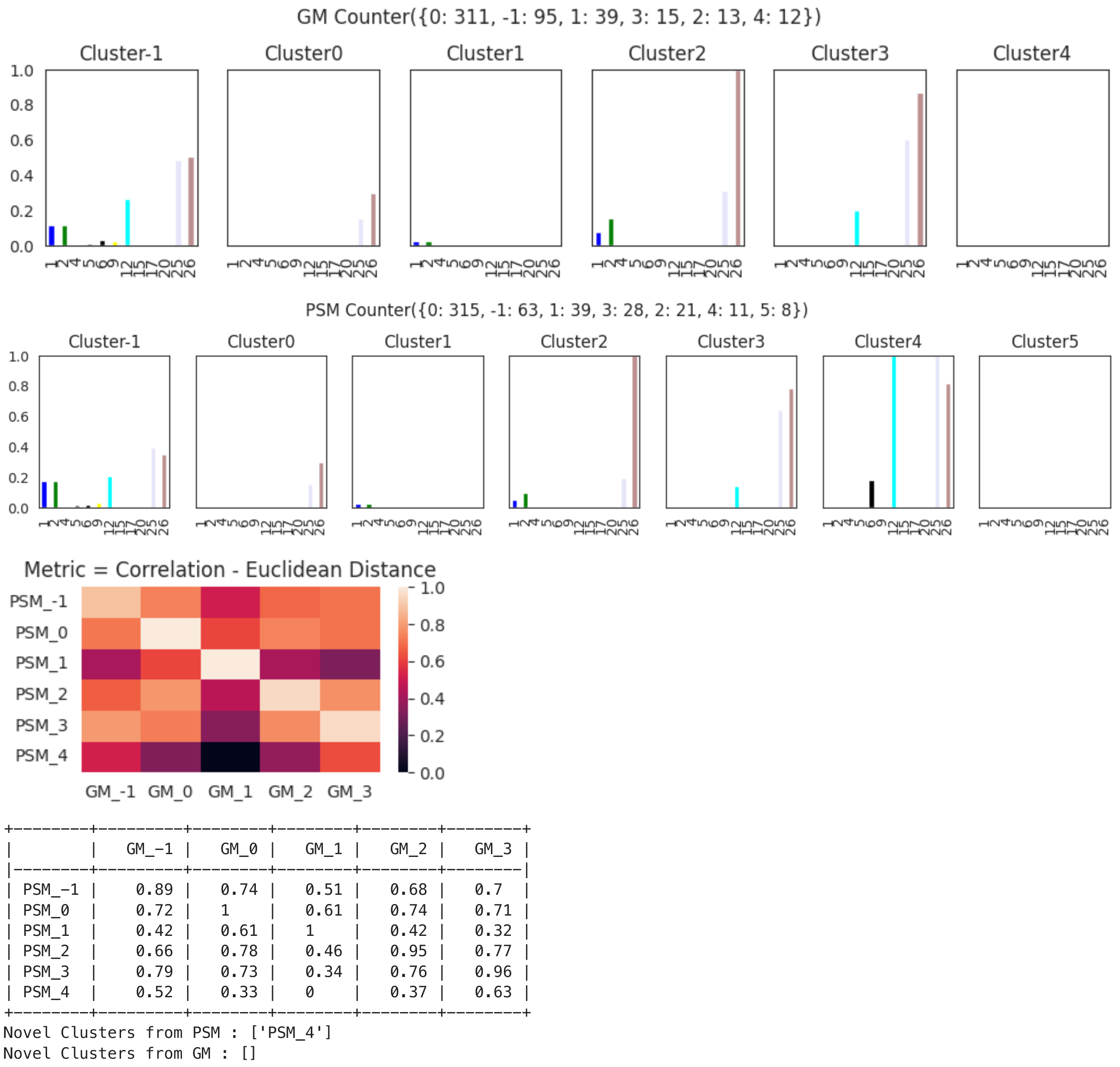}
   \vspace*{-4mm}
   \caption{SN002}
\end{figure*}
\begin{figure*}
\captionsetup{labelformat=empty,labelsep=none}
  \centering
   \includegraphics[width=1\linewidth]{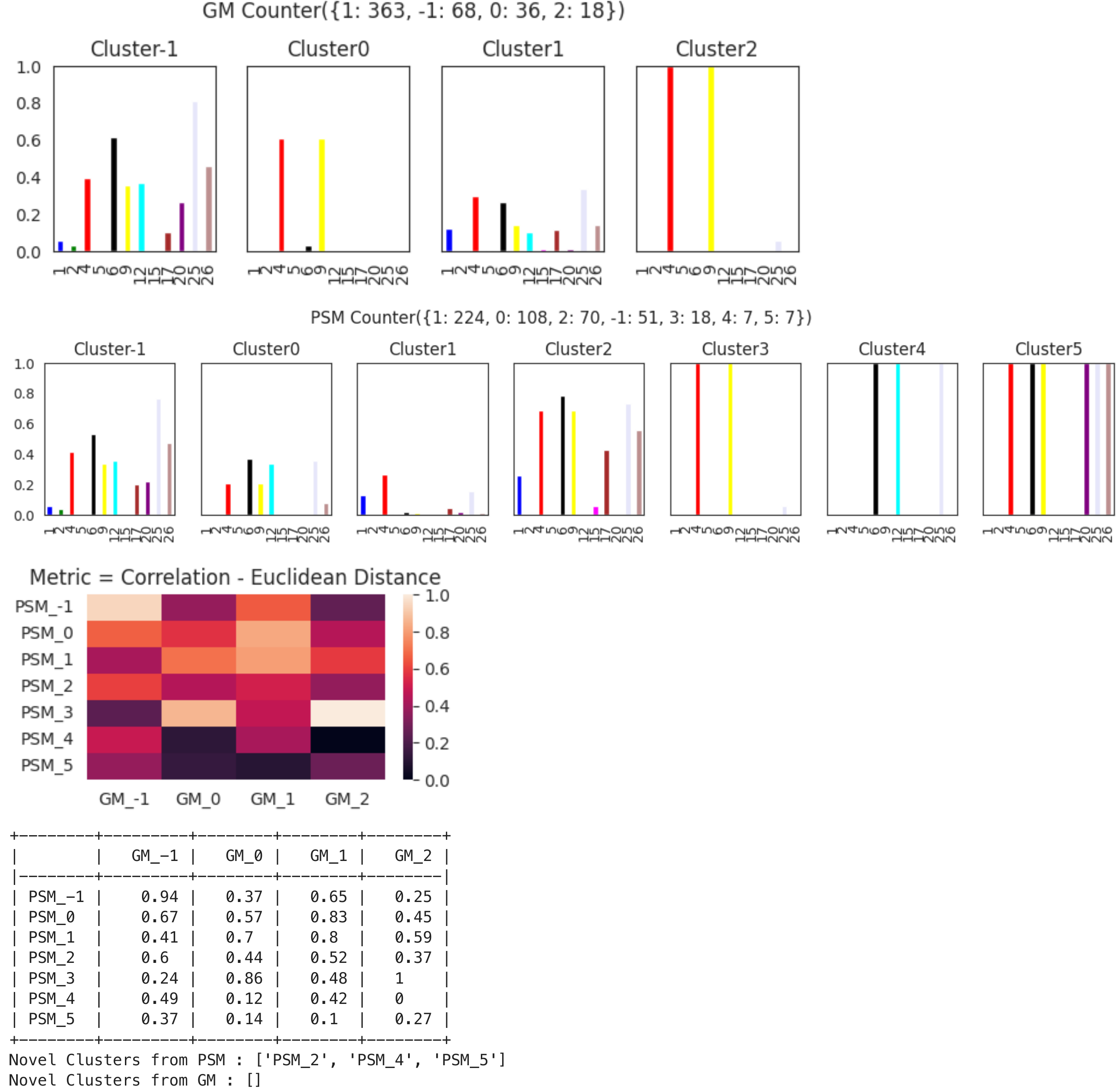}
   \vspace*{-4mm}
   \caption{SN003}
\end{figure*}
\begin{figure*}
\captionsetup{labelformat=empty,labelsep=none}
  \centering
   \includegraphics[width=1\linewidth]{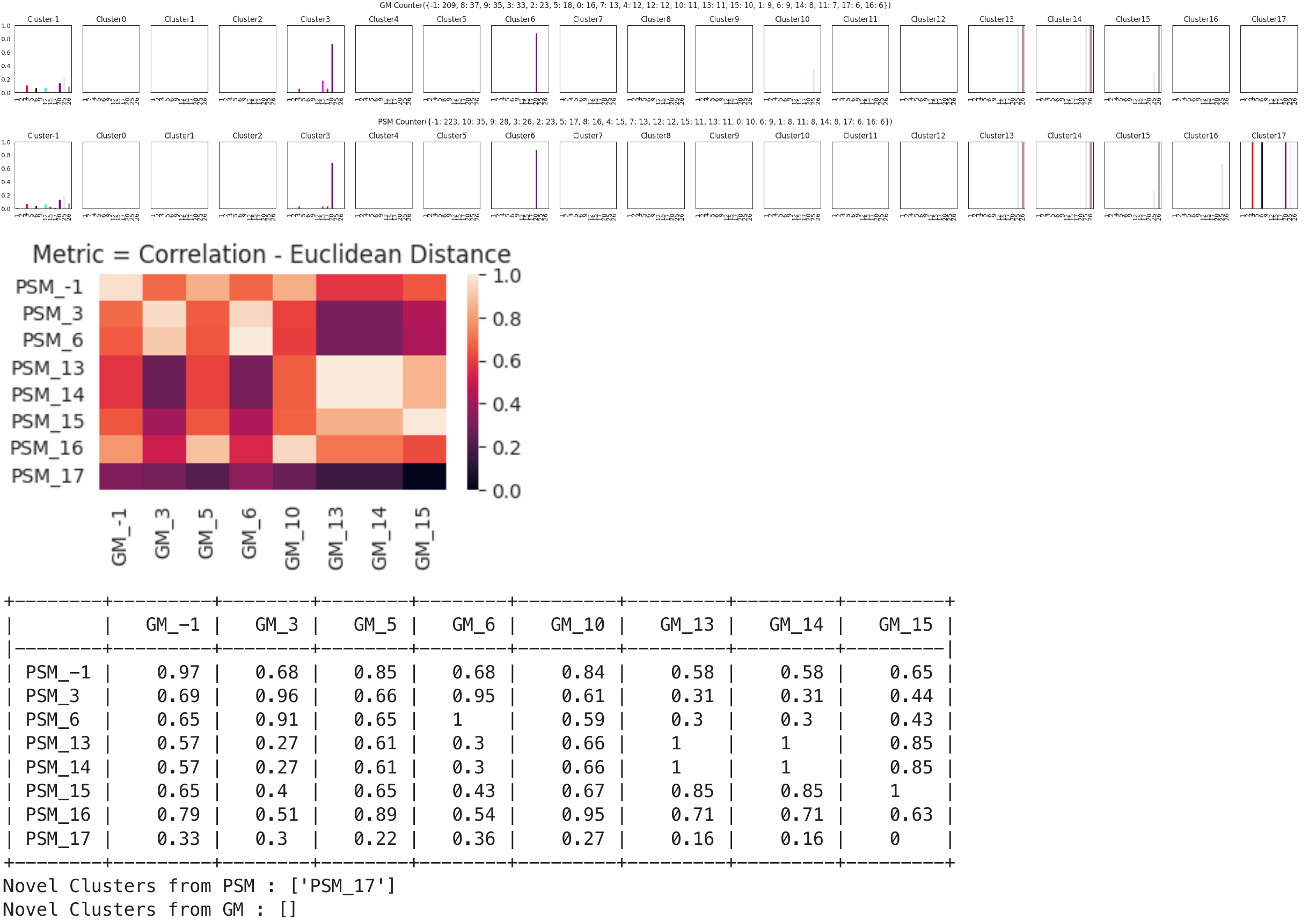}
   \vspace*{-4mm}
   \caption{SN005}
\end{figure*}
\begin{figure*}
\captionsetup{labelformat=empty,labelsep=none}
  \centering
   \includegraphics[width=1\linewidth]{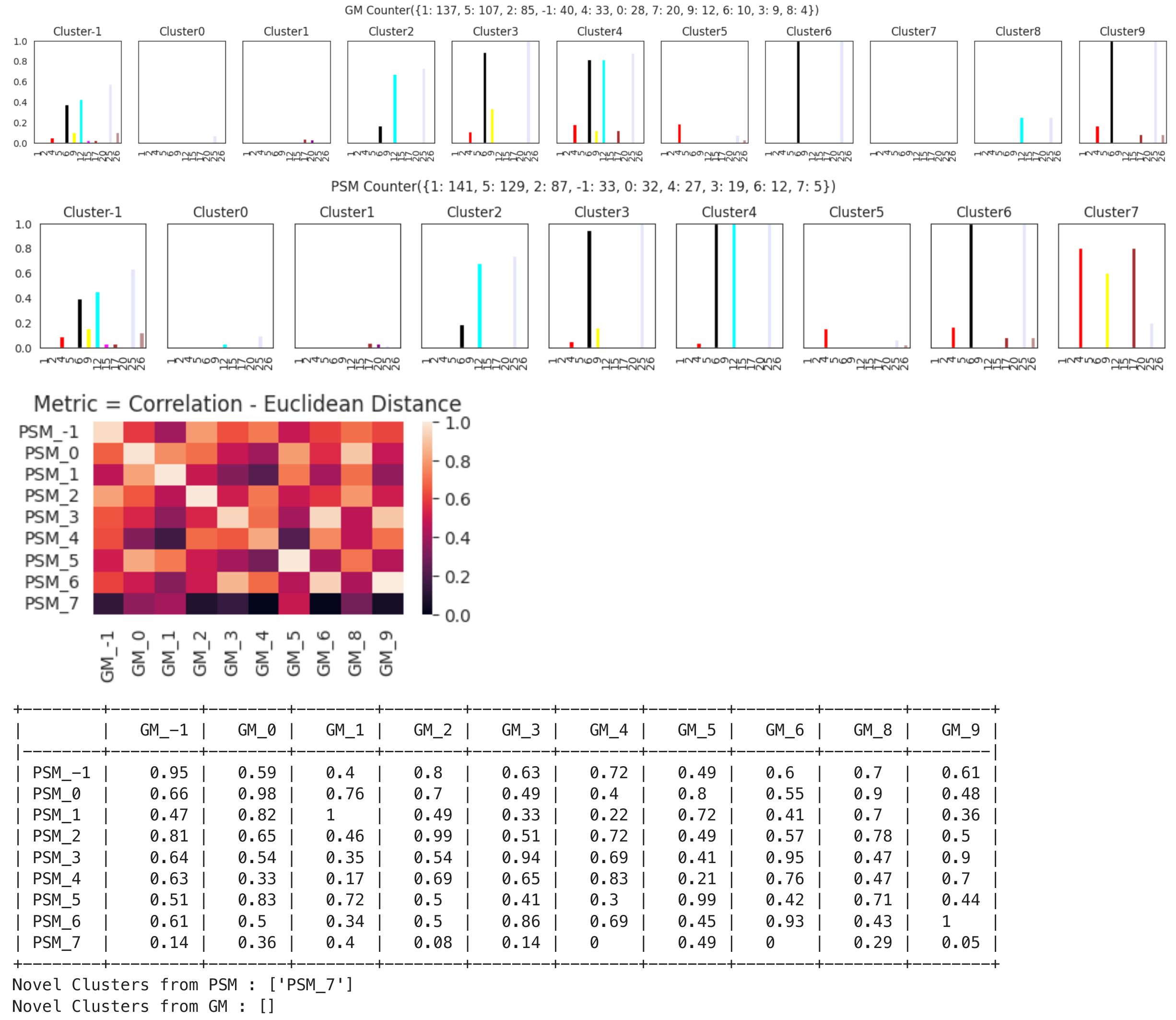}
   \vspace*{-4mm}
   \caption{SN006}
\end{figure*}
\begin{figure*}
\captionsetup{labelformat=empty,labelsep=none}
  \centering
   \includegraphics[width=1\linewidth]{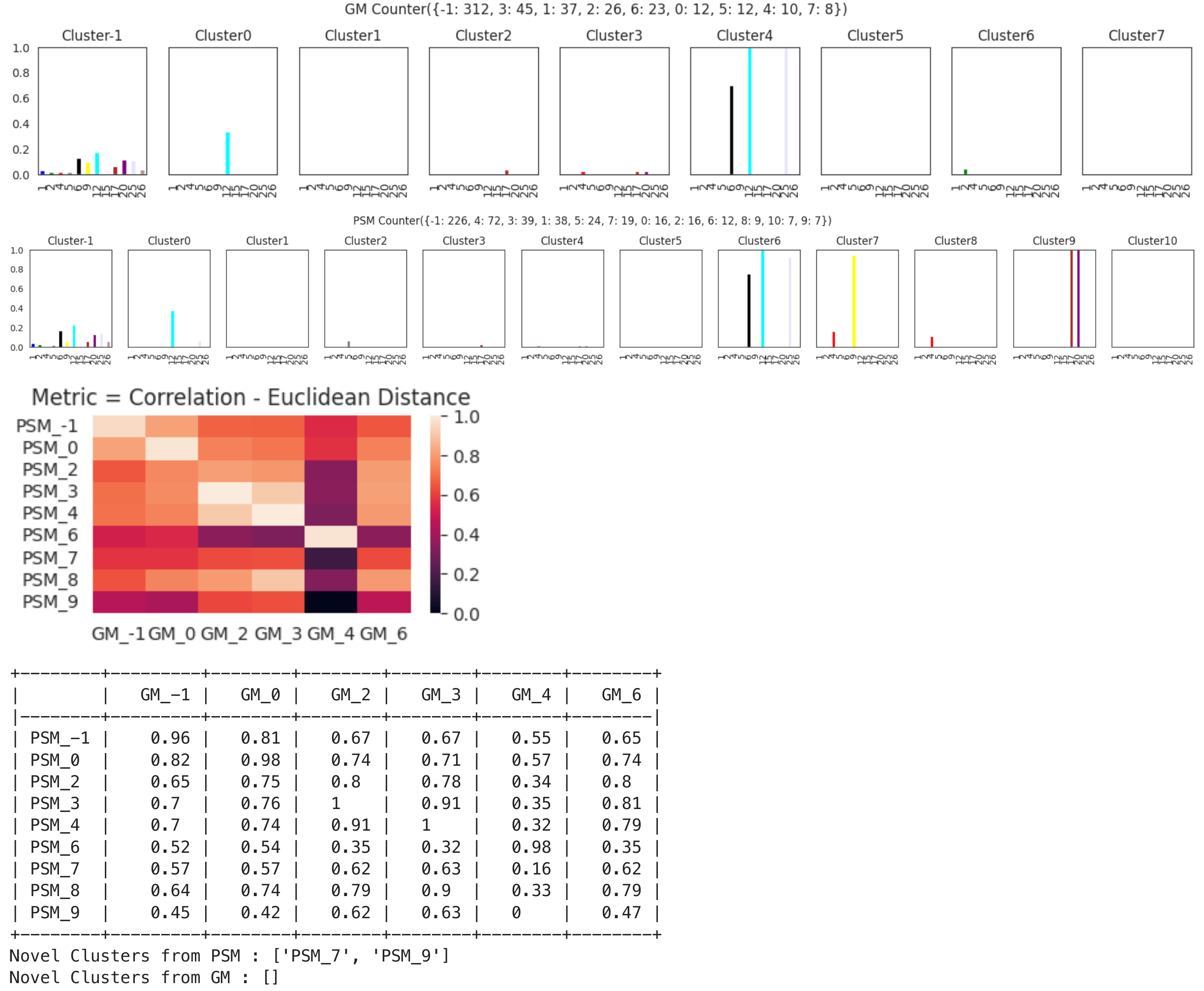}
   \vspace*{-4mm}
   \caption{SN007}
\end{figure*}
\begin{figure*}
\captionsetup{labelformat=empty,labelsep=none}
  \centering
   \includegraphics[width=1\linewidth]{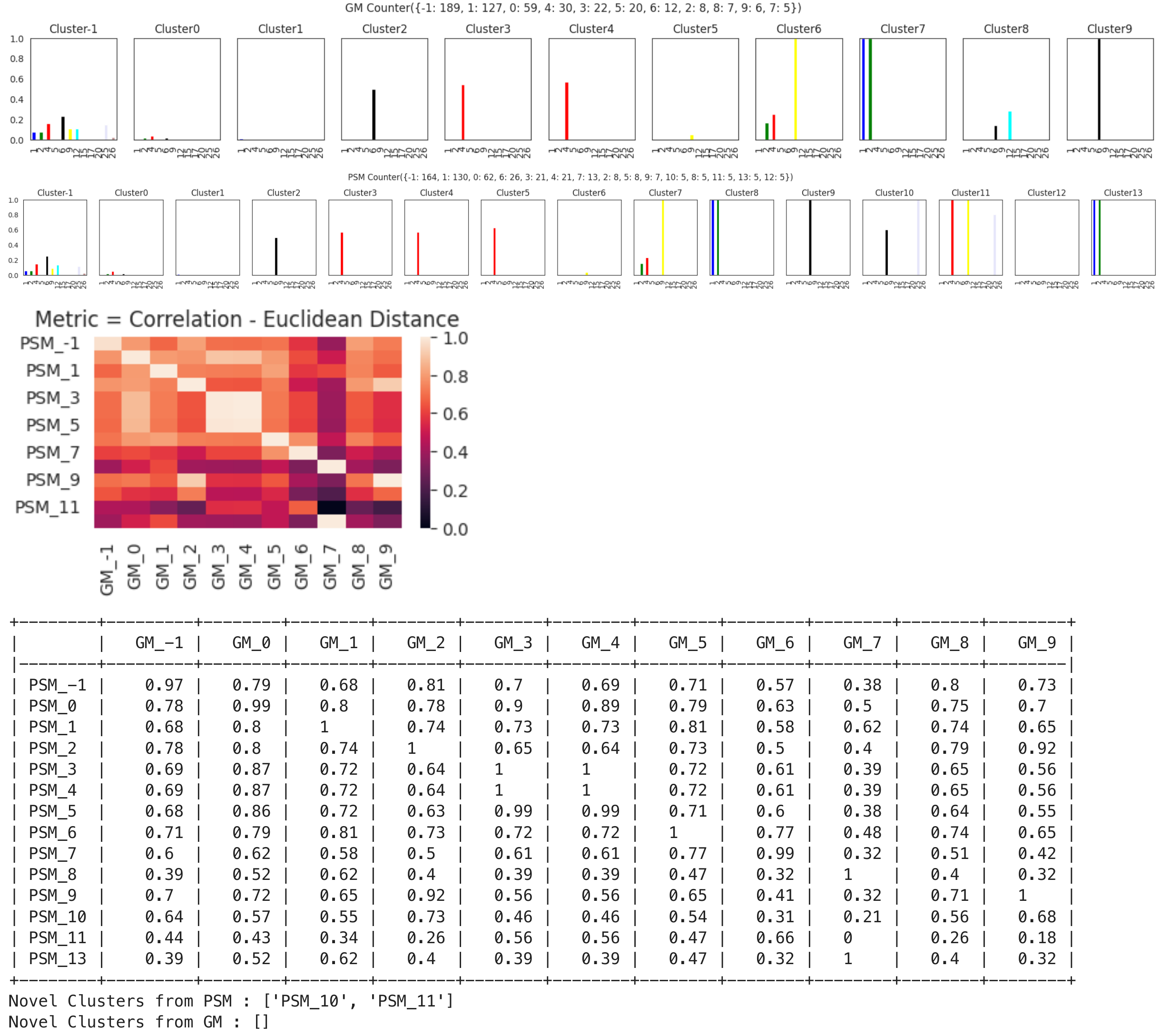}
   \vspace*{-4mm}
   \caption{SN008}
\end{figure*}
\begin{figure*}
\captionsetup{labelformat=empty,labelsep=none}
  \centering
   \includegraphics[width=1\linewidth]{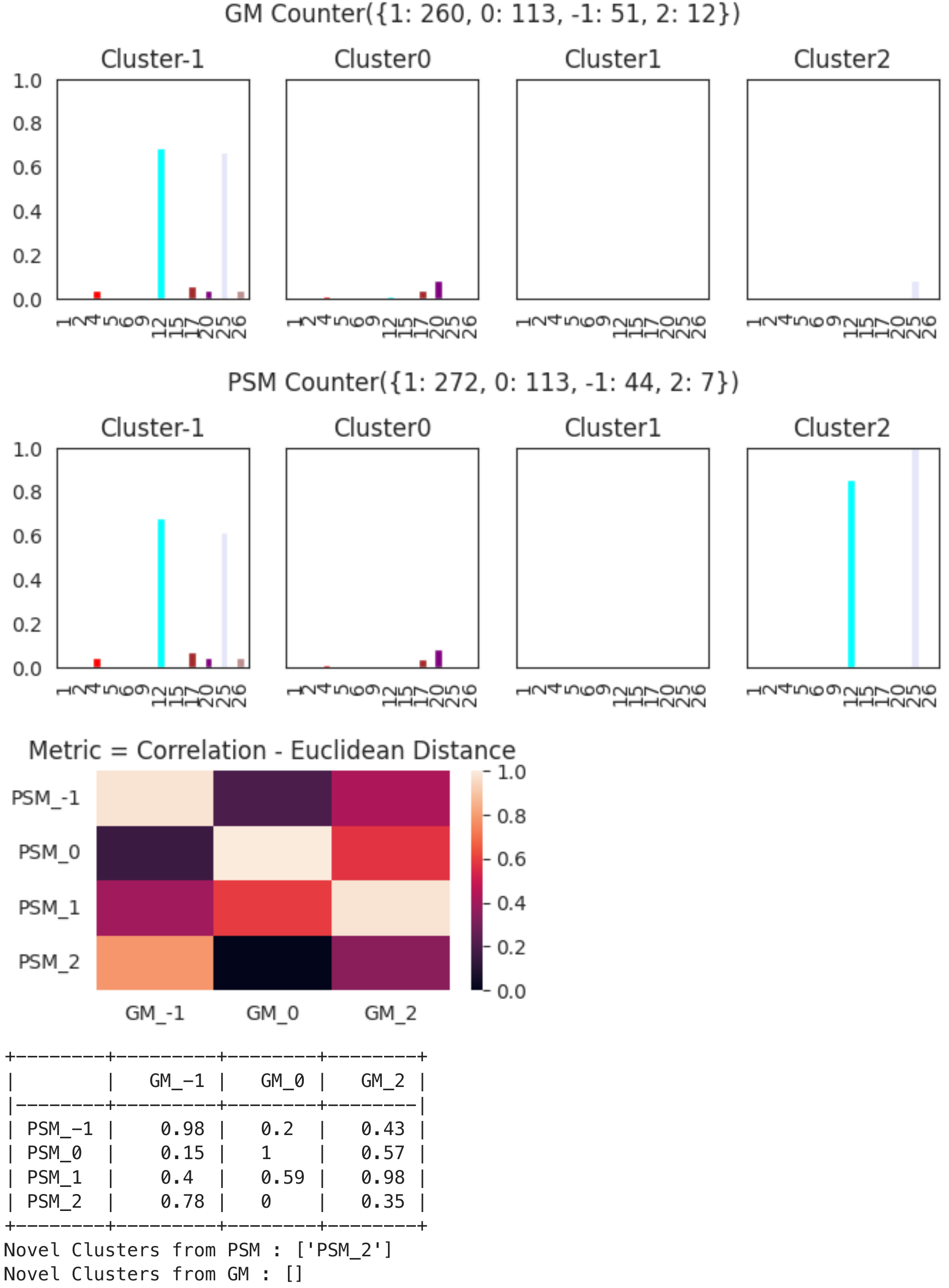}
   \vspace*{-4mm}
   \caption{SN009}
\end{figure*}
\begin{figure*}
\captionsetup{labelformat=empty,labelsep=none}
  \centering
   \includegraphics[width=1\linewidth]{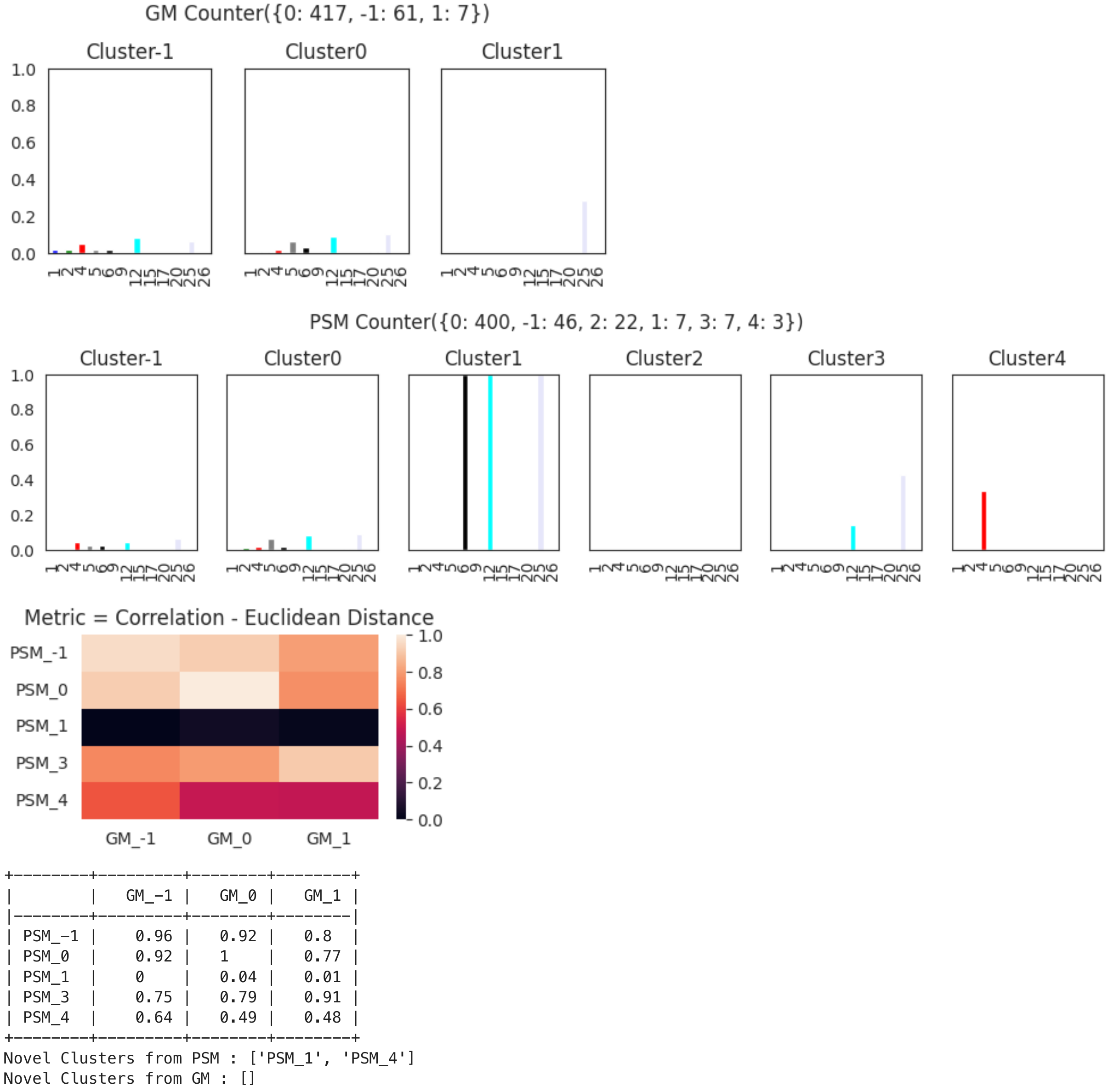}
   \vspace*{-4mm}
   \caption{SN010}
\end{figure*}
\begin{figure*}
\captionsetup{labelformat=empty,labelsep=none}
  \centering
   \includegraphics[width=1\linewidth]{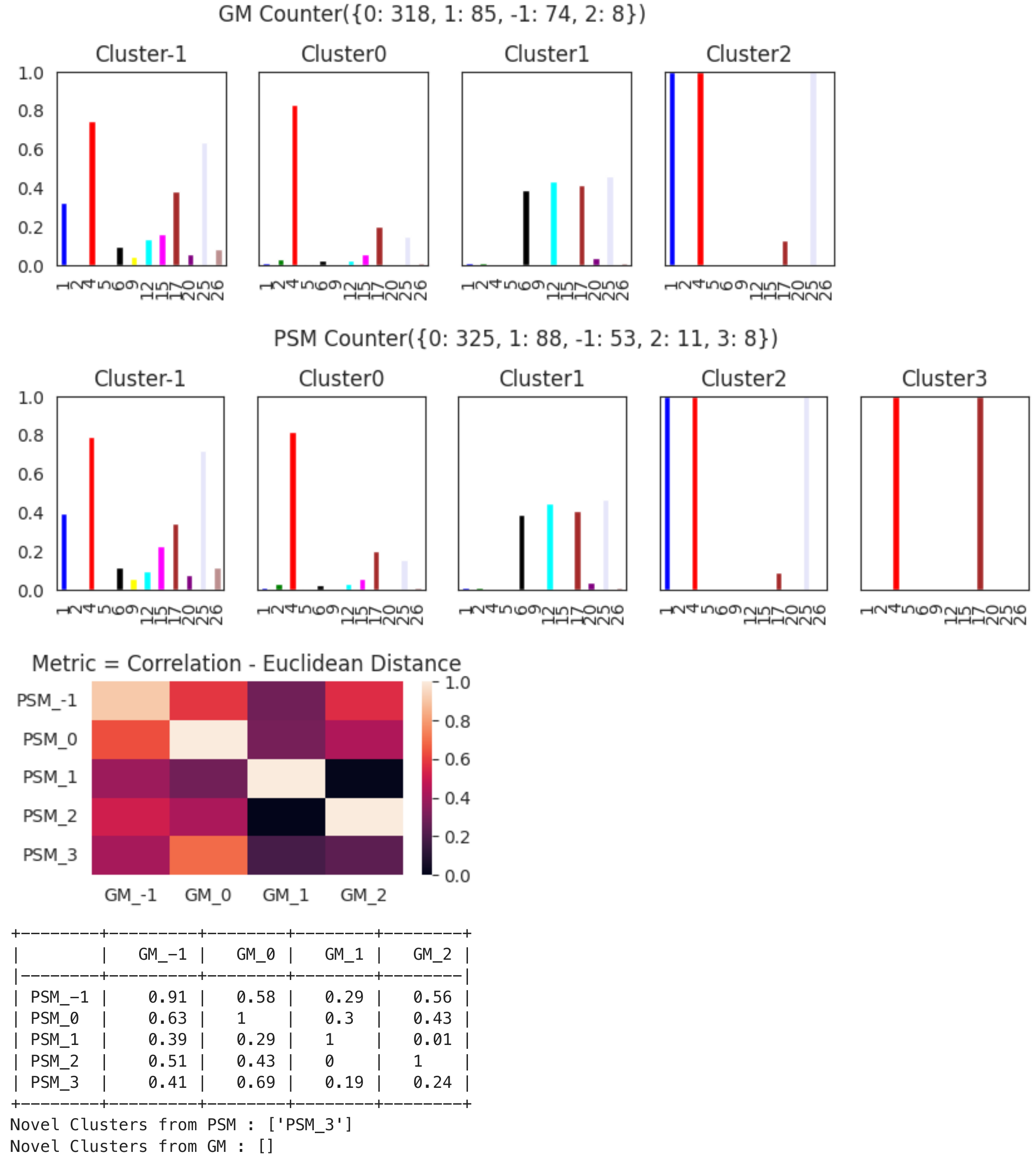}
   \vspace*{-4mm}
   \caption{SN011}
\end{figure*}
\begin{figure*}
\captionsetup{labelformat=empty,labelsep=none}
  \centering
   \includegraphics[width=1\linewidth]{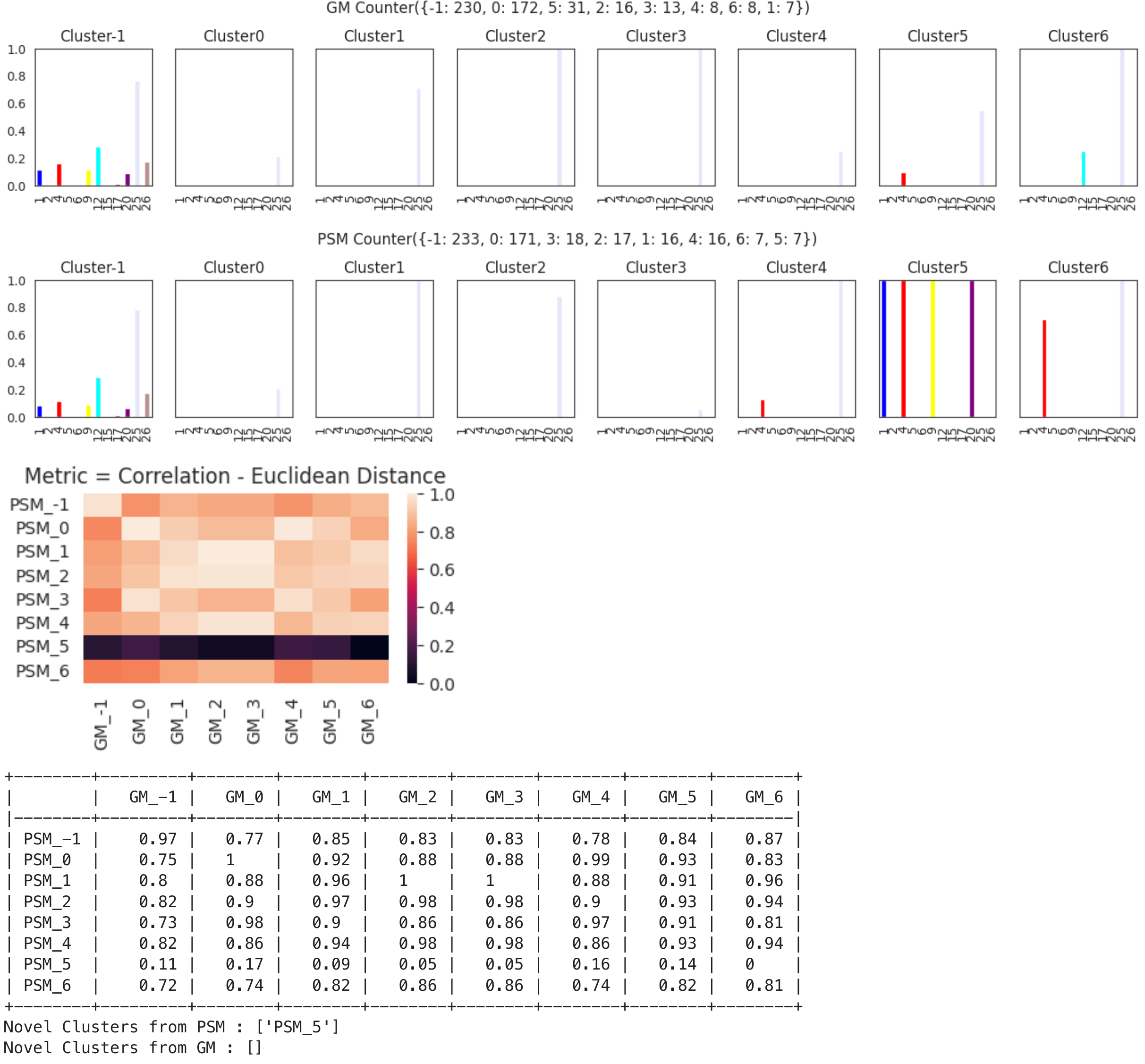}
   \vspace*{-4mm}
   \caption{SN012}
\end{figure*}
\begin{figure*}
\captionsetup{labelformat=empty,labelsep=none}
  \centering
   \includegraphics[width=1\linewidth]{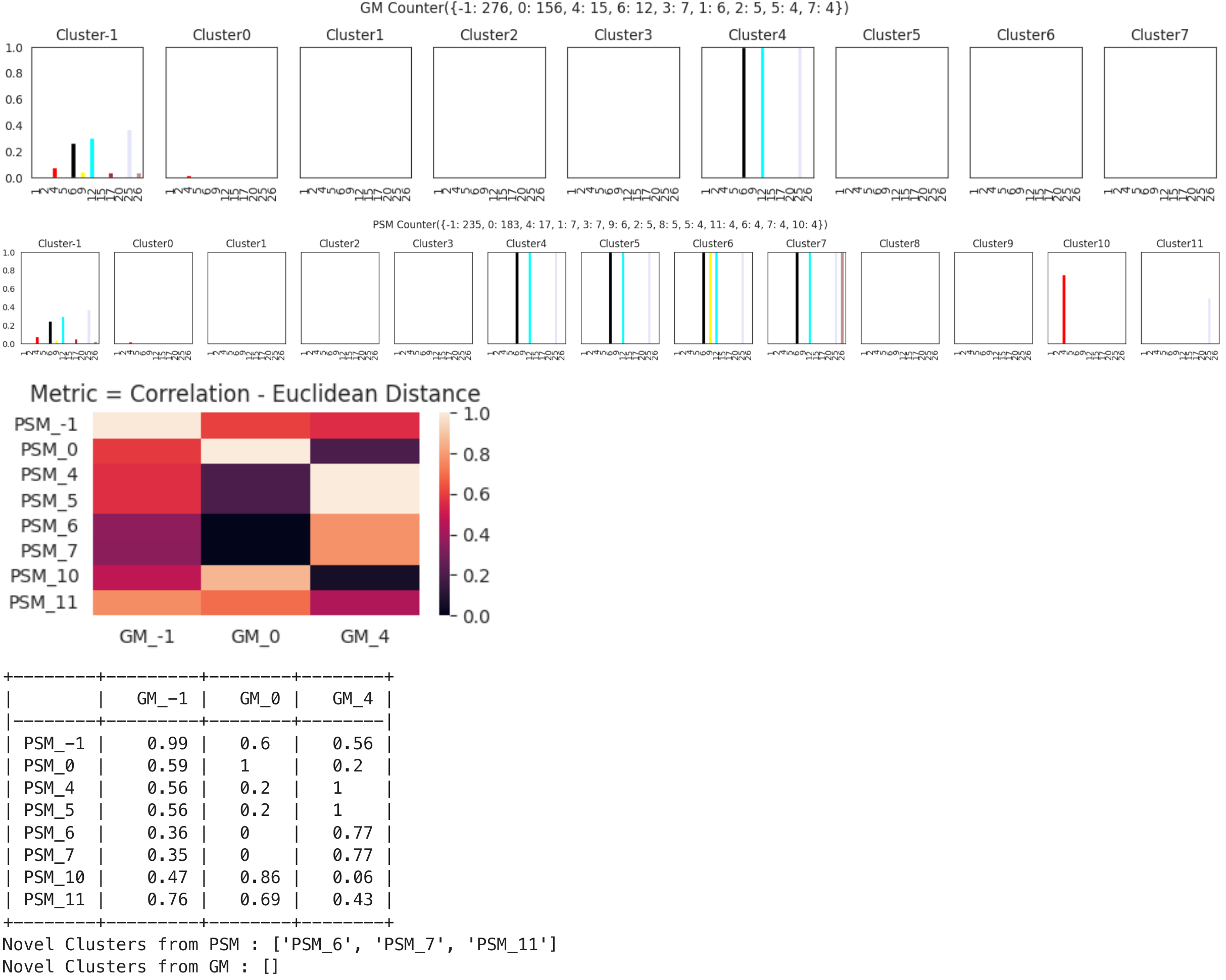}
   \vspace*{-4mm}
   \caption{SN013}
\end{figure*}
\begin{figure*}
\captionsetup{labelformat=empty,labelsep=none}
  \centering
   \includegraphics[width=1\linewidth]{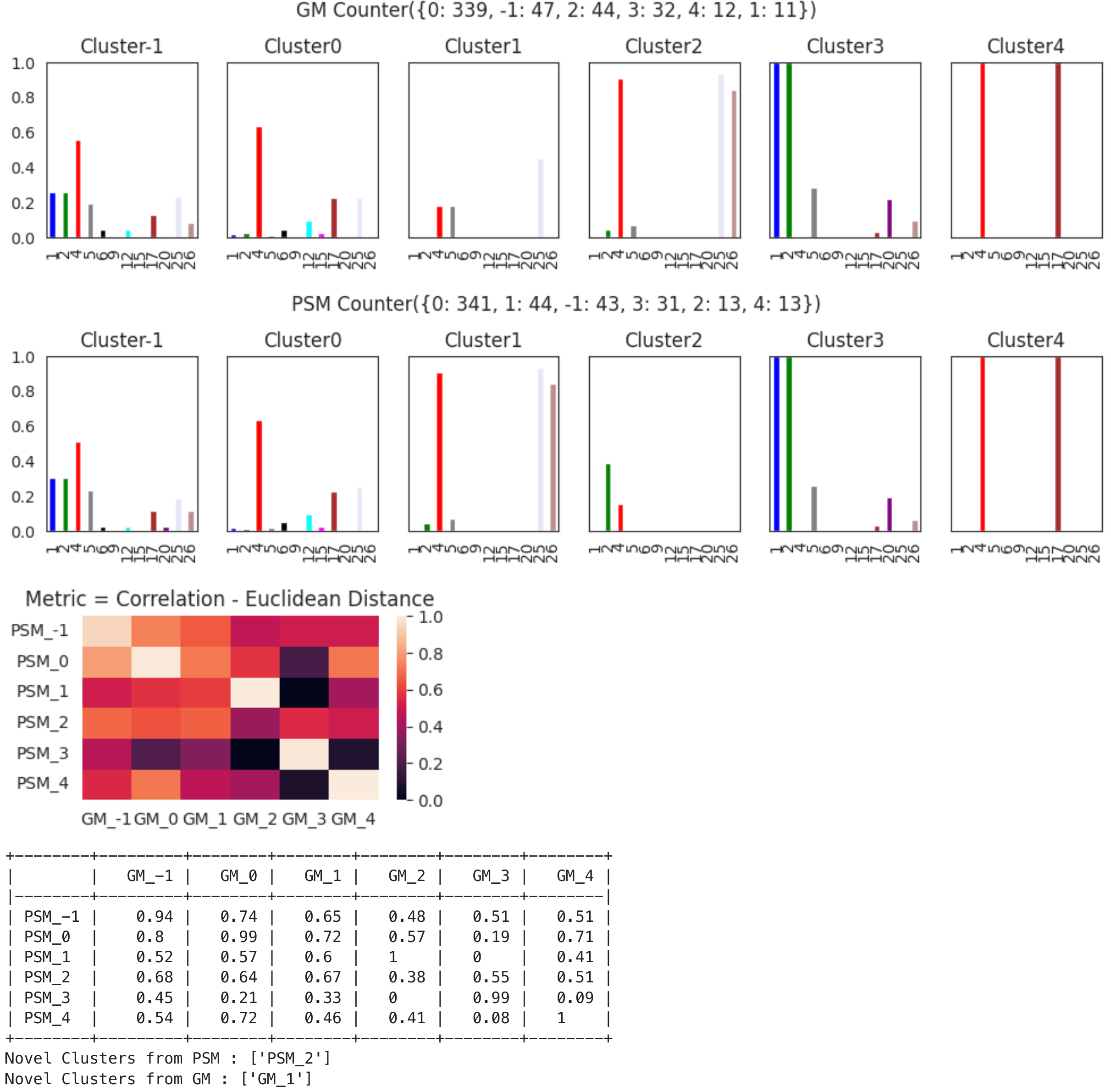}
   \vspace*{-4mm}
   \caption{SN016}
\end{figure*}
\begin{figure*}
\captionsetup{labelformat=empty,labelsep=none}
  \centering
   \includegraphics[width=1\linewidth]{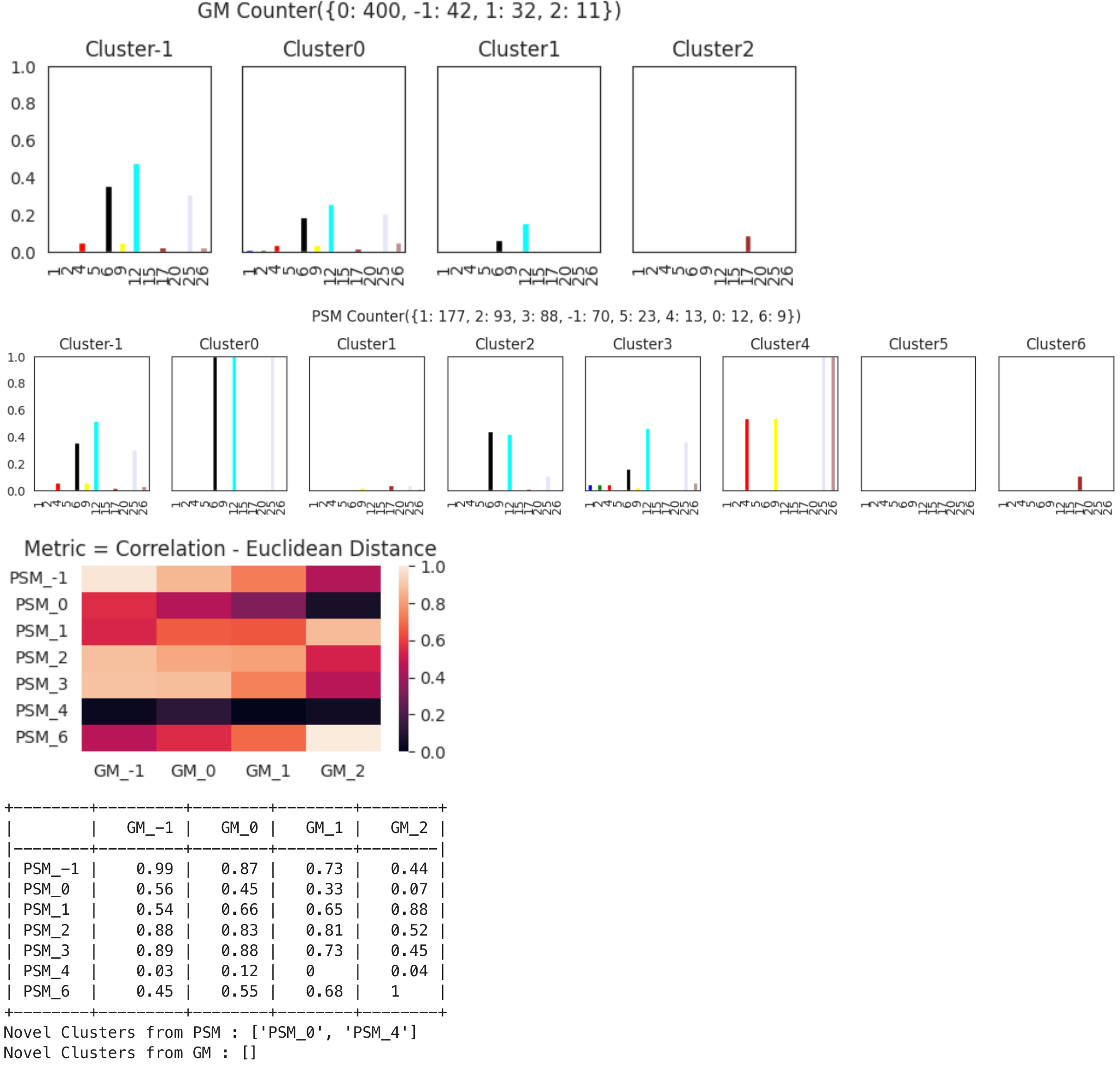}
   \vspace*{-4mm}
   \caption{SN017}
\end{figure*}
\begin{figure*}
\captionsetup{labelformat=empty,labelsep=none}
  \centering
   \includegraphics[width=1\linewidth]{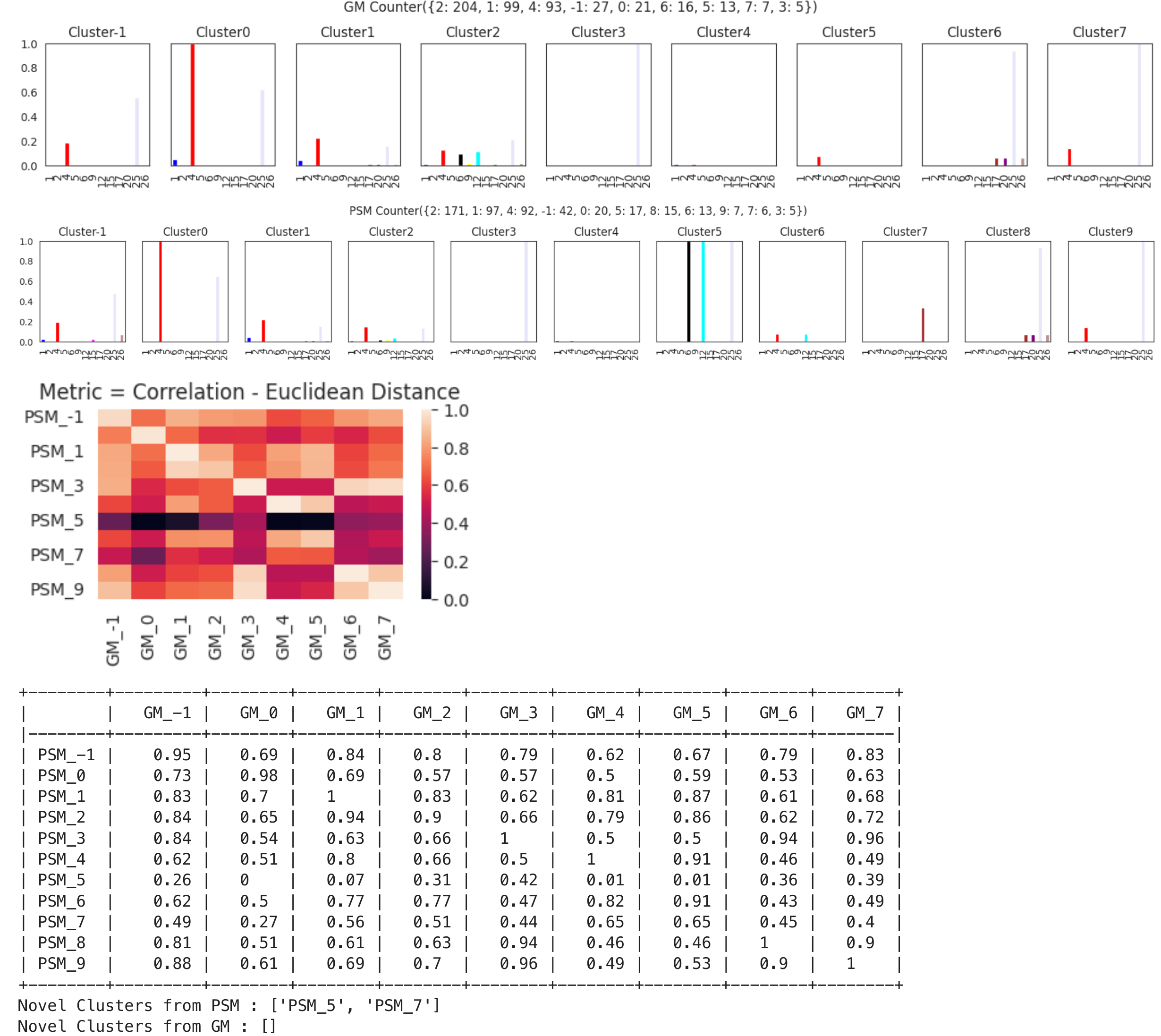}
   \vspace*{-4mm}
   \caption{SN021}
\end{figure*}
\begin{figure*}
\captionsetup{labelformat=empty,labelsep=none}
  \centering
   \includegraphics[width=1\linewidth]{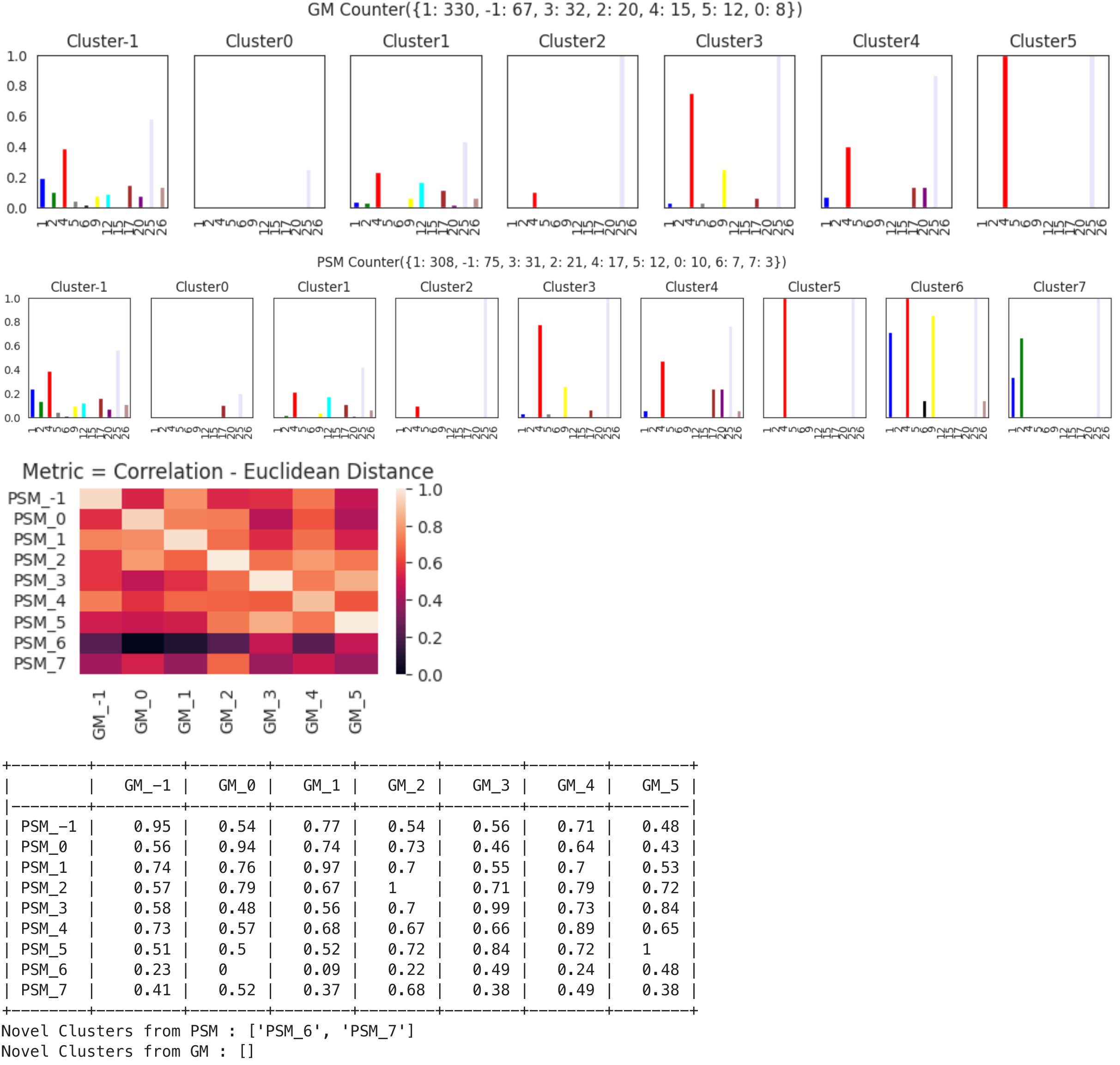}
   \vspace*{-4mm}
   \caption{SN023}
\end{figure*}
\begin{figure*}
\captionsetup{labelformat=empty,labelsep=none}
  \centering
   \includegraphics[width=1\linewidth]{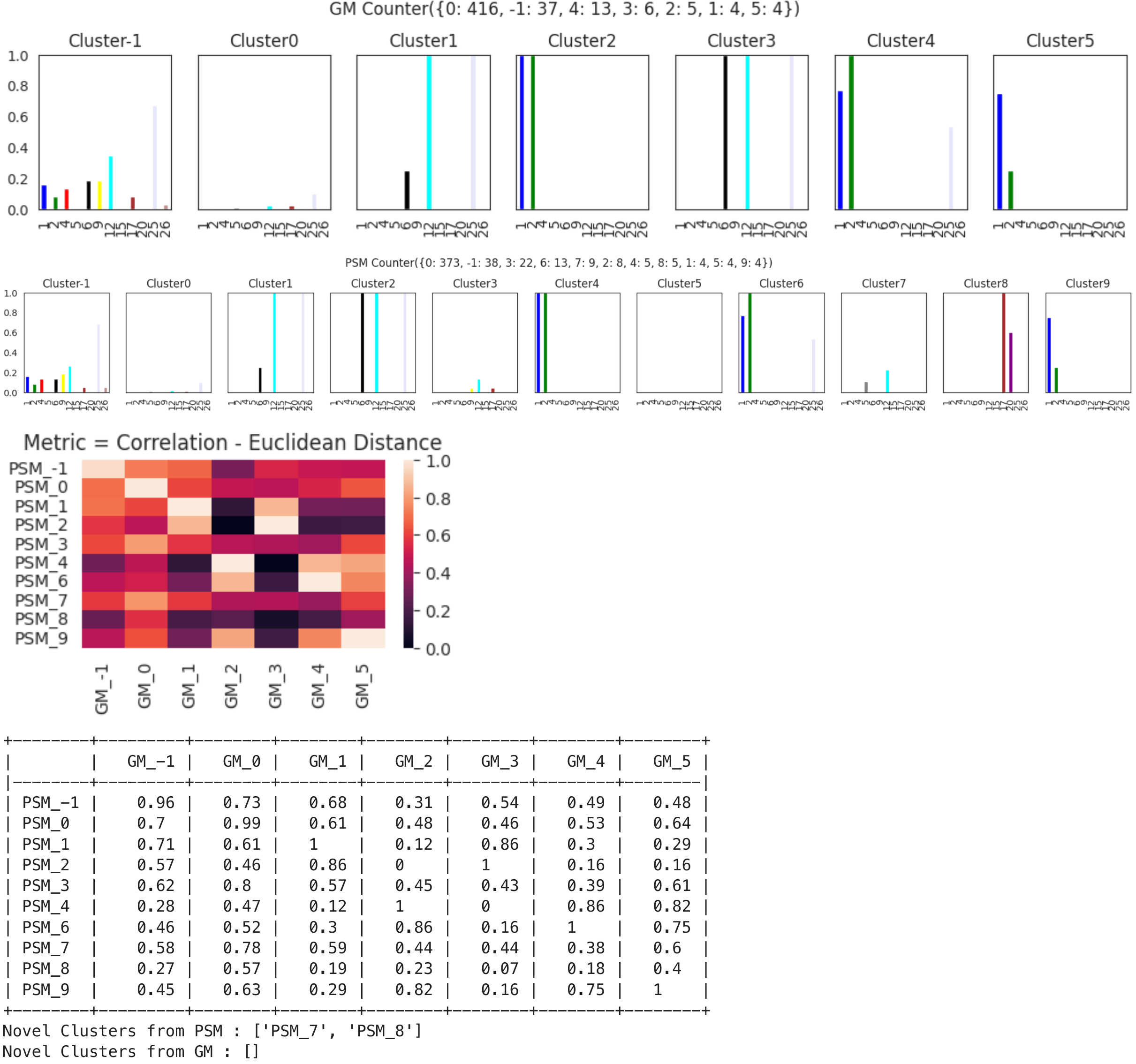}
   \vspace*{-4mm}
   \caption{SN024}
\end{figure*}
\begin{figure*}
\captionsetup{labelformat=empty,labelsep=none}
  \centering
   \includegraphics[width=1\linewidth]{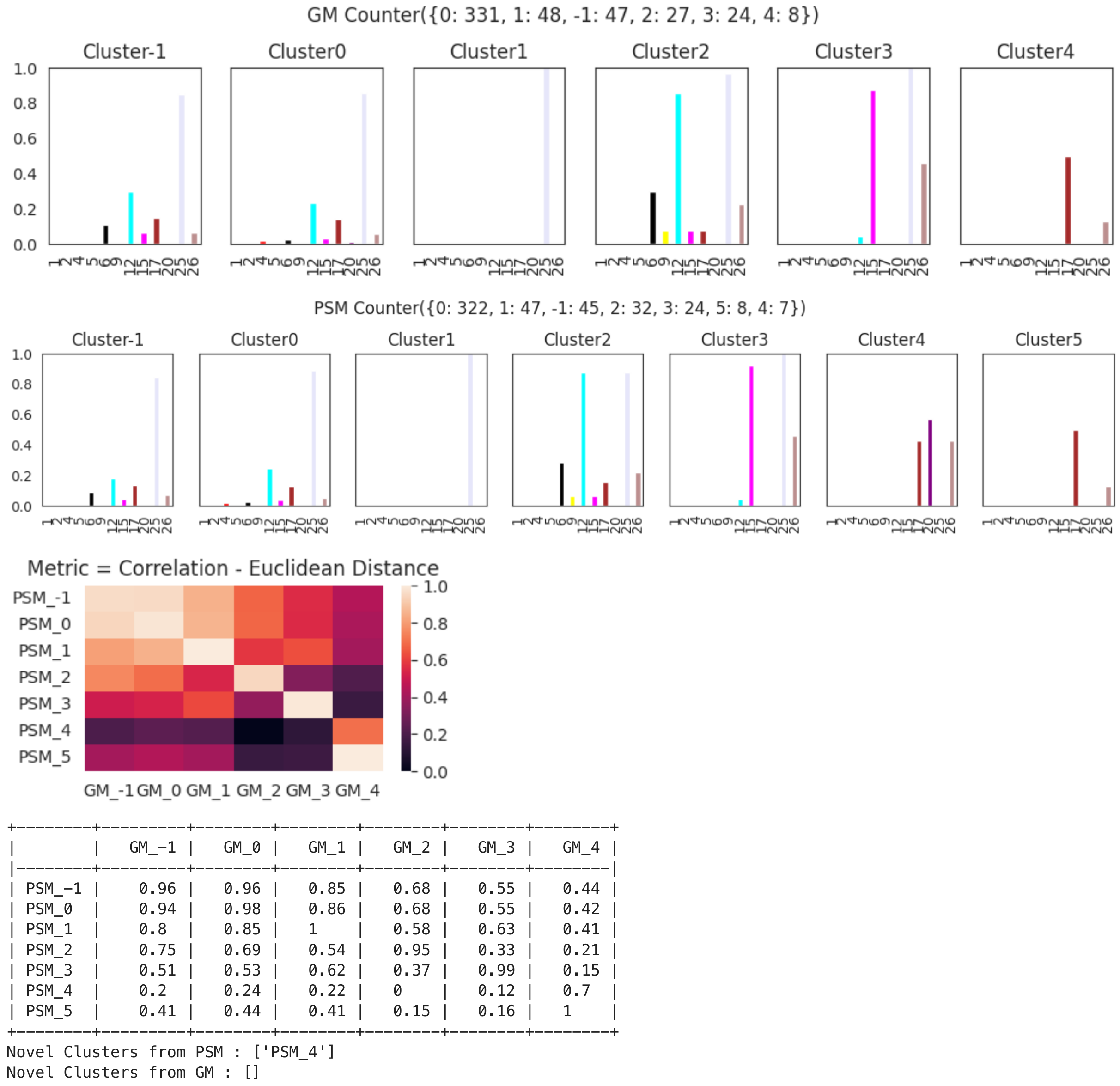}
   \vspace*{-4mm}
   \caption{SN025}
\end{figure*}
\begin{figure*}
\captionsetup{labelformat=empty,labelsep=none}
  \centering
   \includegraphics[width=1\linewidth]{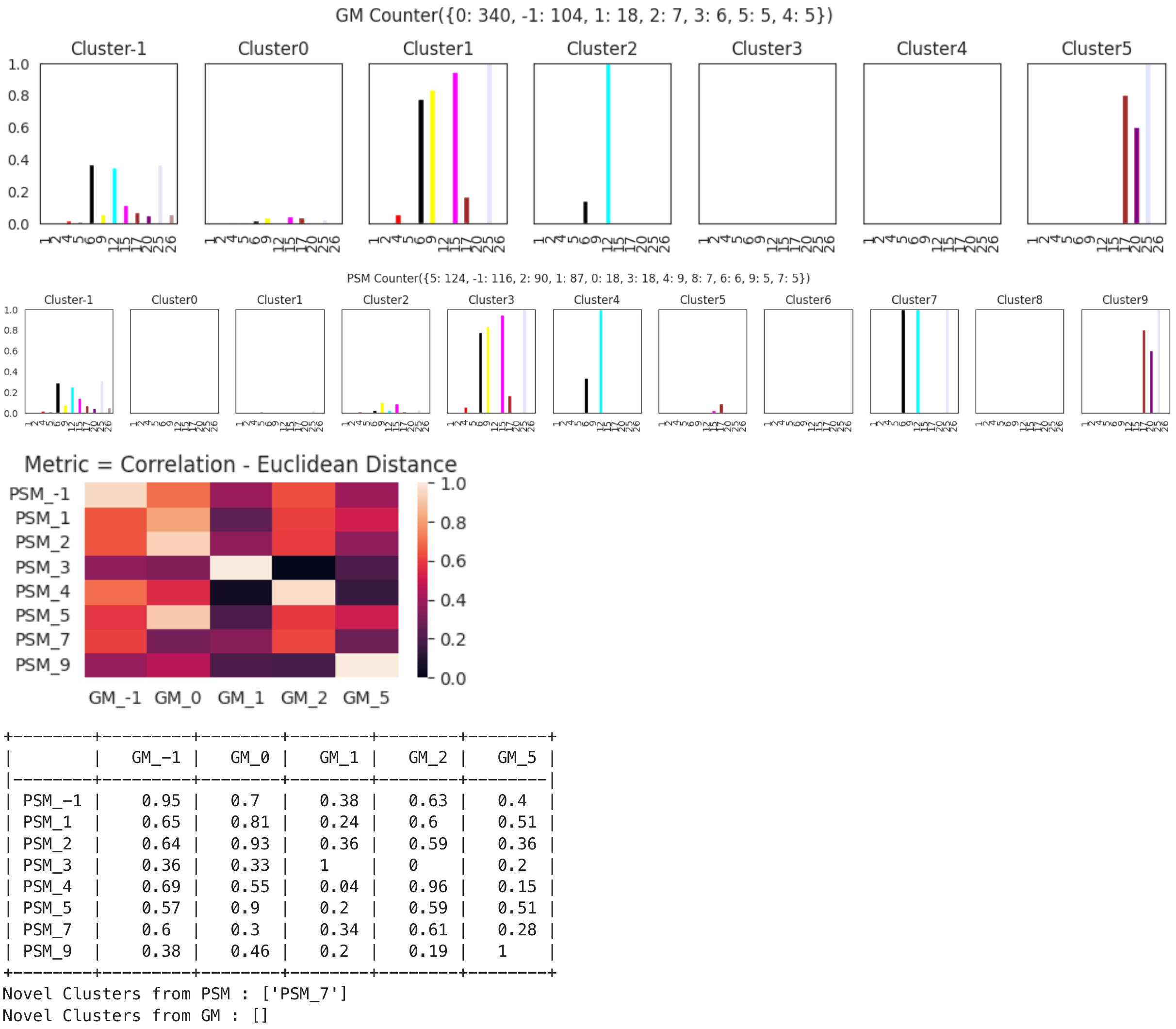}
   \vspace*{-4mm}
   \caption{SN026}
\end{figure*}
\begin{figure*}
\captionsetup{labelformat=empty,labelsep=none}
  \centering
   \includegraphics[width=1\linewidth]{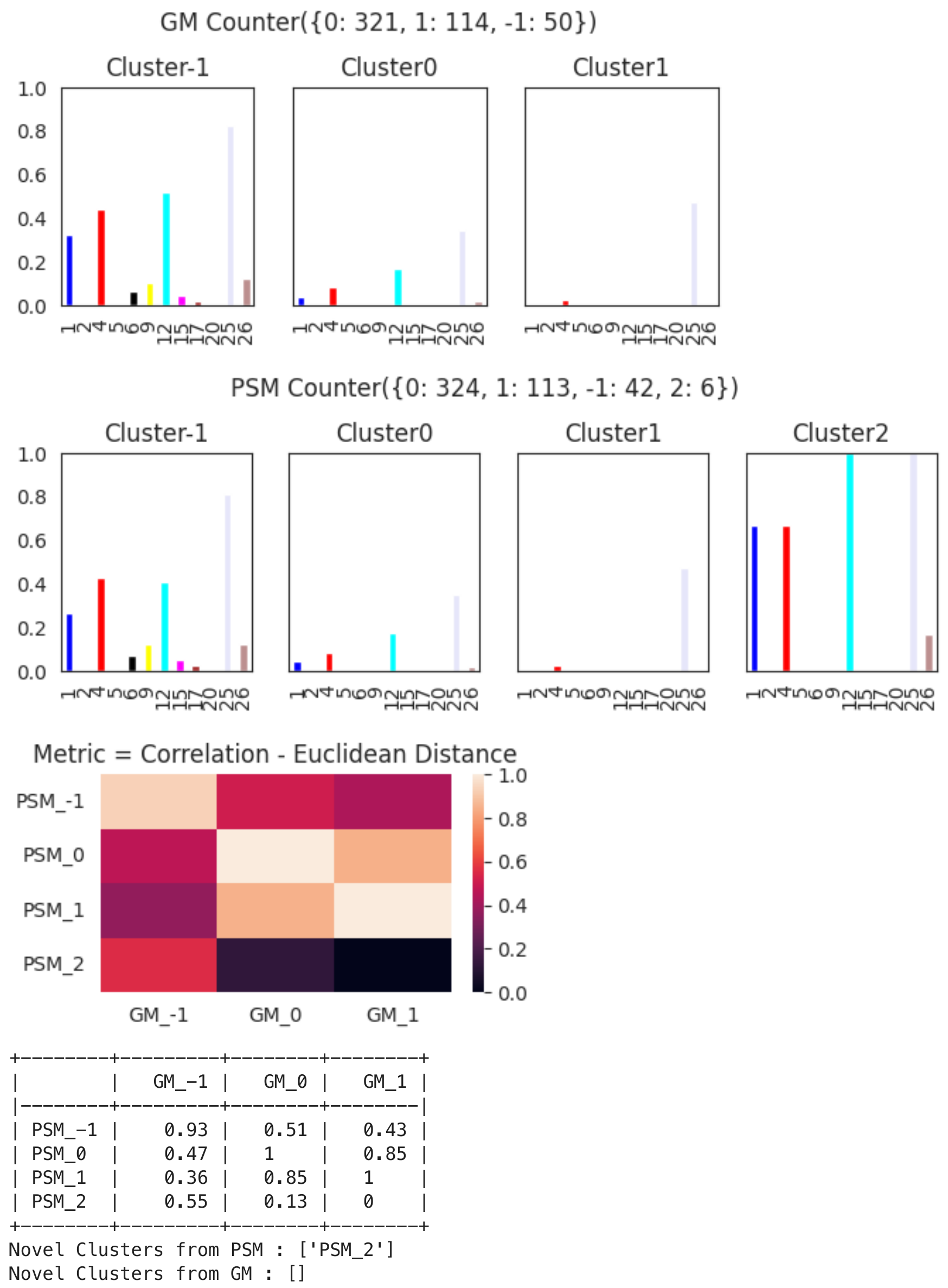}
   \vspace*{-4mm}
   \caption{SN027}
\end{figure*}
\begin{figure*}
\captionsetup{labelformat=empty,labelsep=none}
  \centering
   \includegraphics[width=1\linewidth]{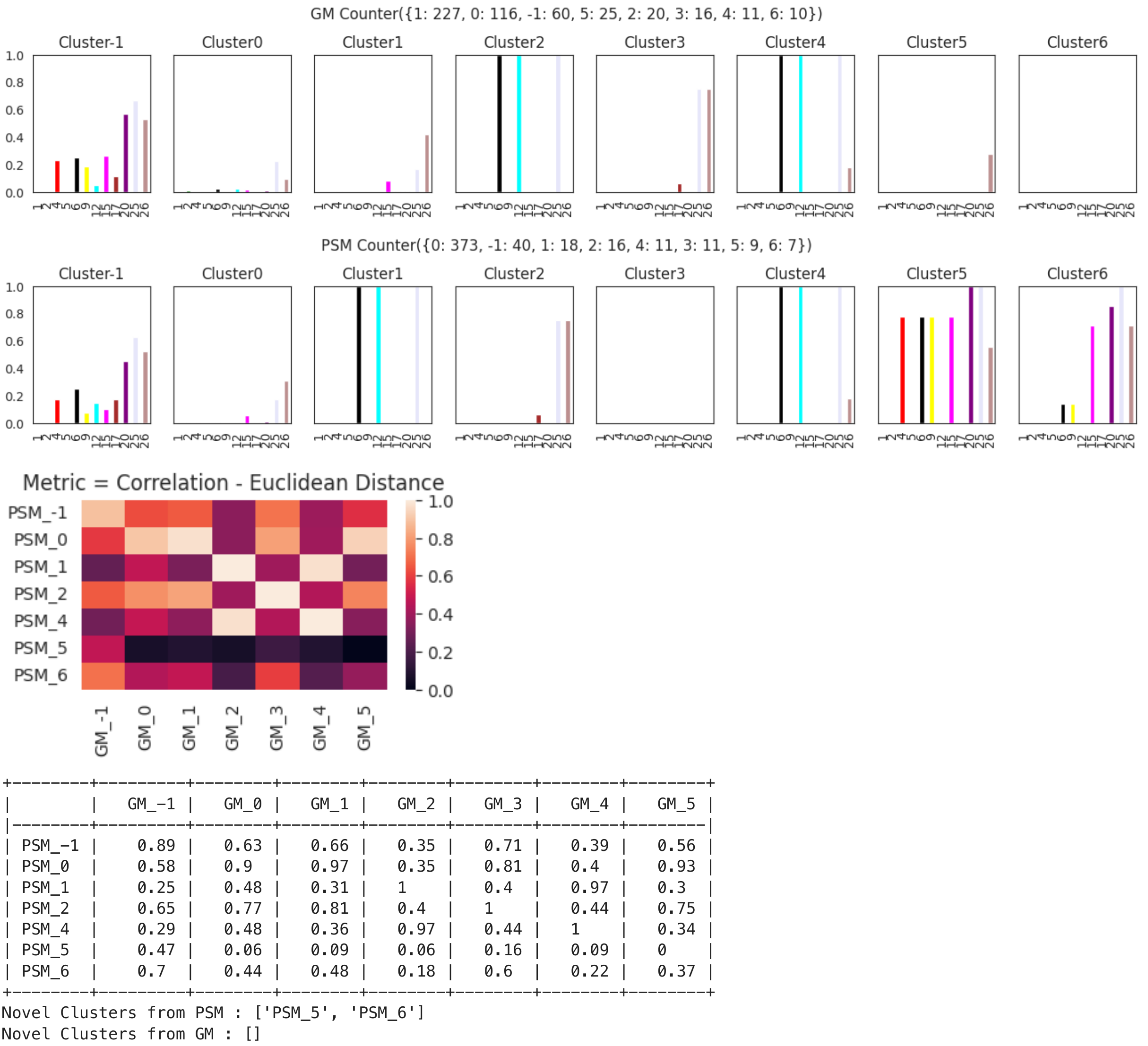}
   \vspace*{-4mm}
   \caption{SN028}
\end{figure*}
\begin{figure*}
\captionsetup{labelformat=empty,labelsep=none}
  \centering
   \includegraphics[width=1\linewidth]{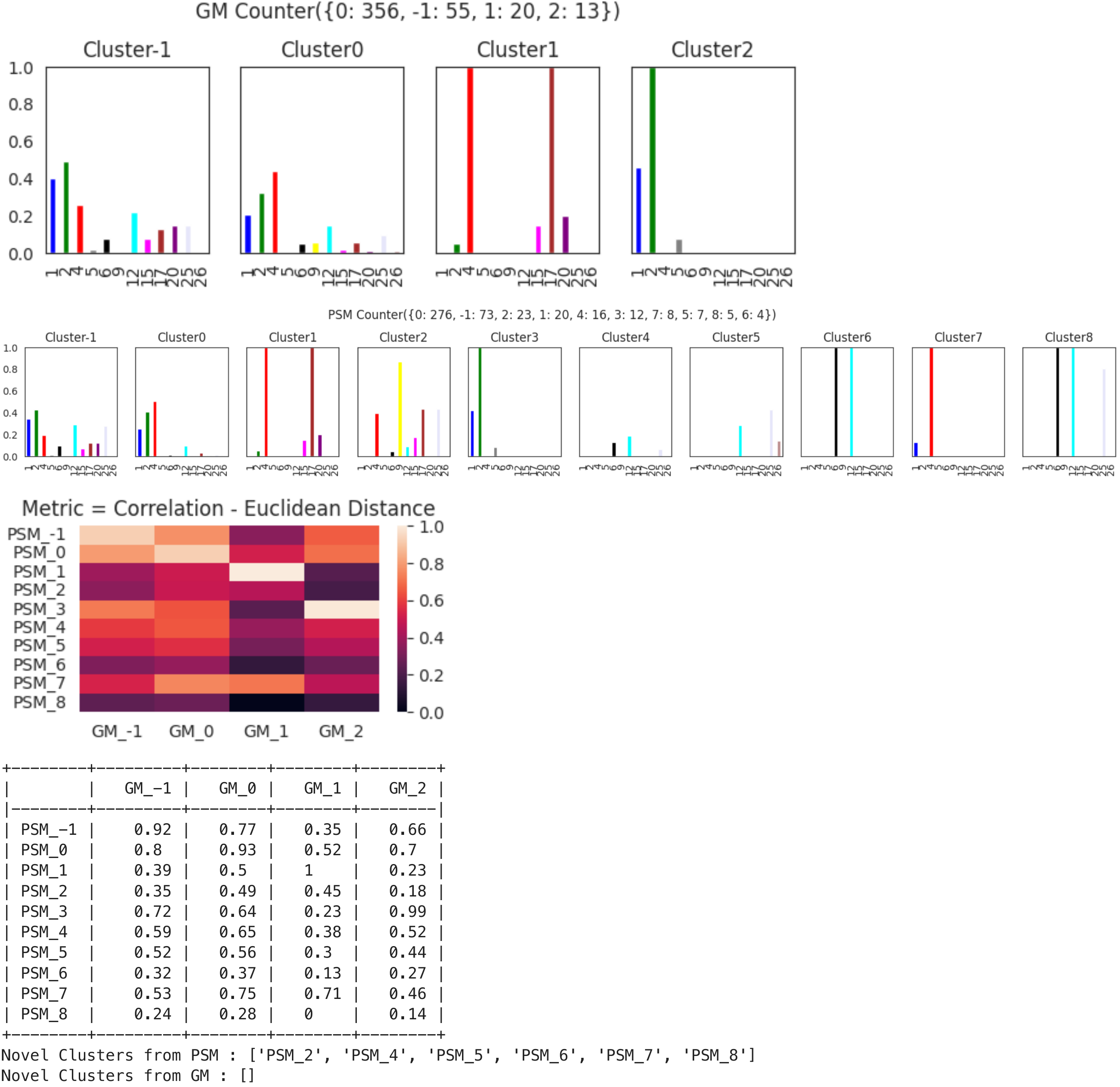}
   \vspace*{-4mm}
   \caption{SN029}
\end{figure*}
\begin{figure*}
\captionsetup{labelformat=empty,labelsep=none}
  \centering
   \includegraphics[width=1\linewidth]{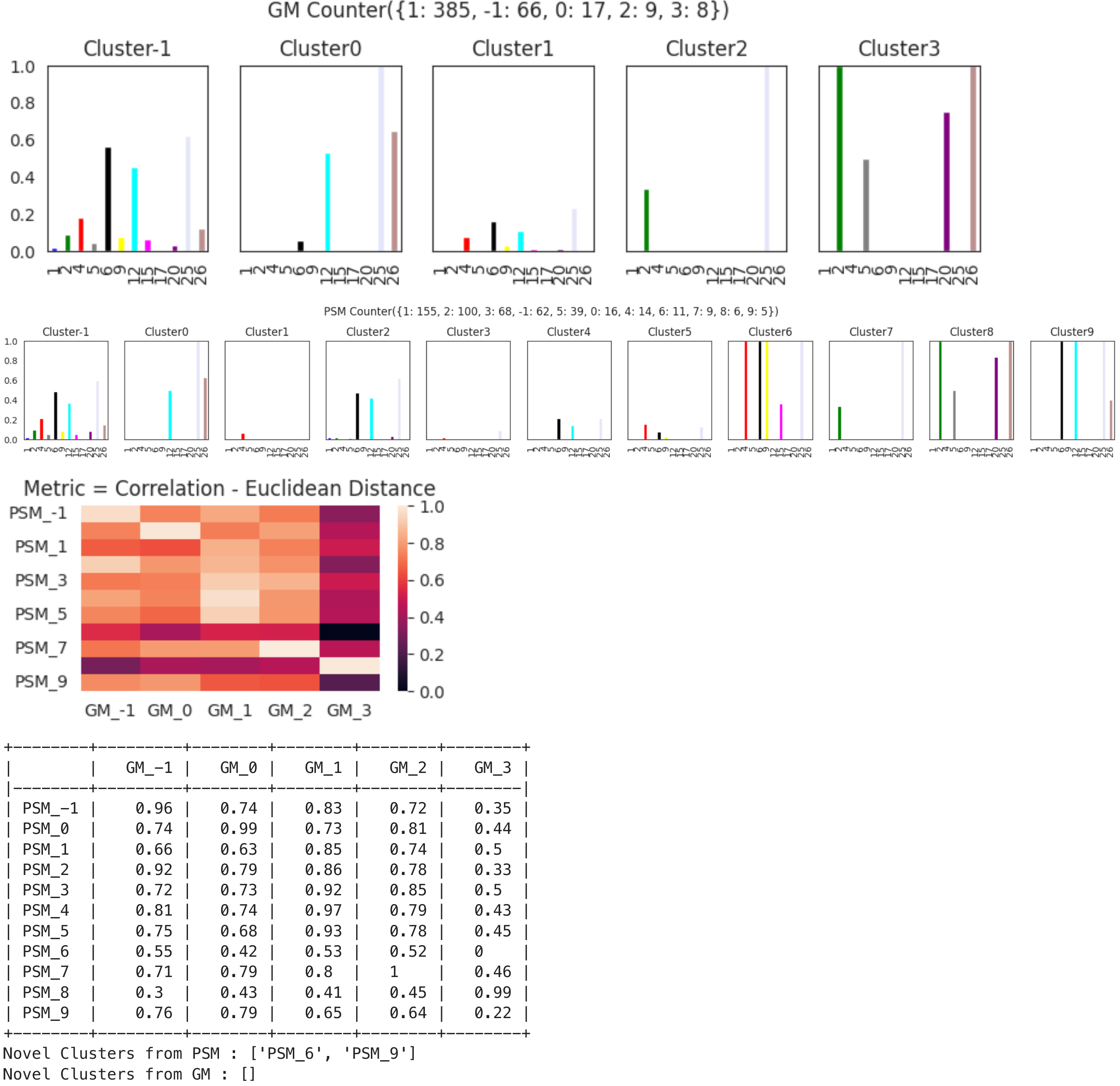}
   \vspace*{-4mm}
   \caption{SN030}
\end{figure*}
\begin{figure*}
\captionsetup{labelformat=empty,labelsep=none}
  \centering
   \includegraphics[width=1\linewidth]{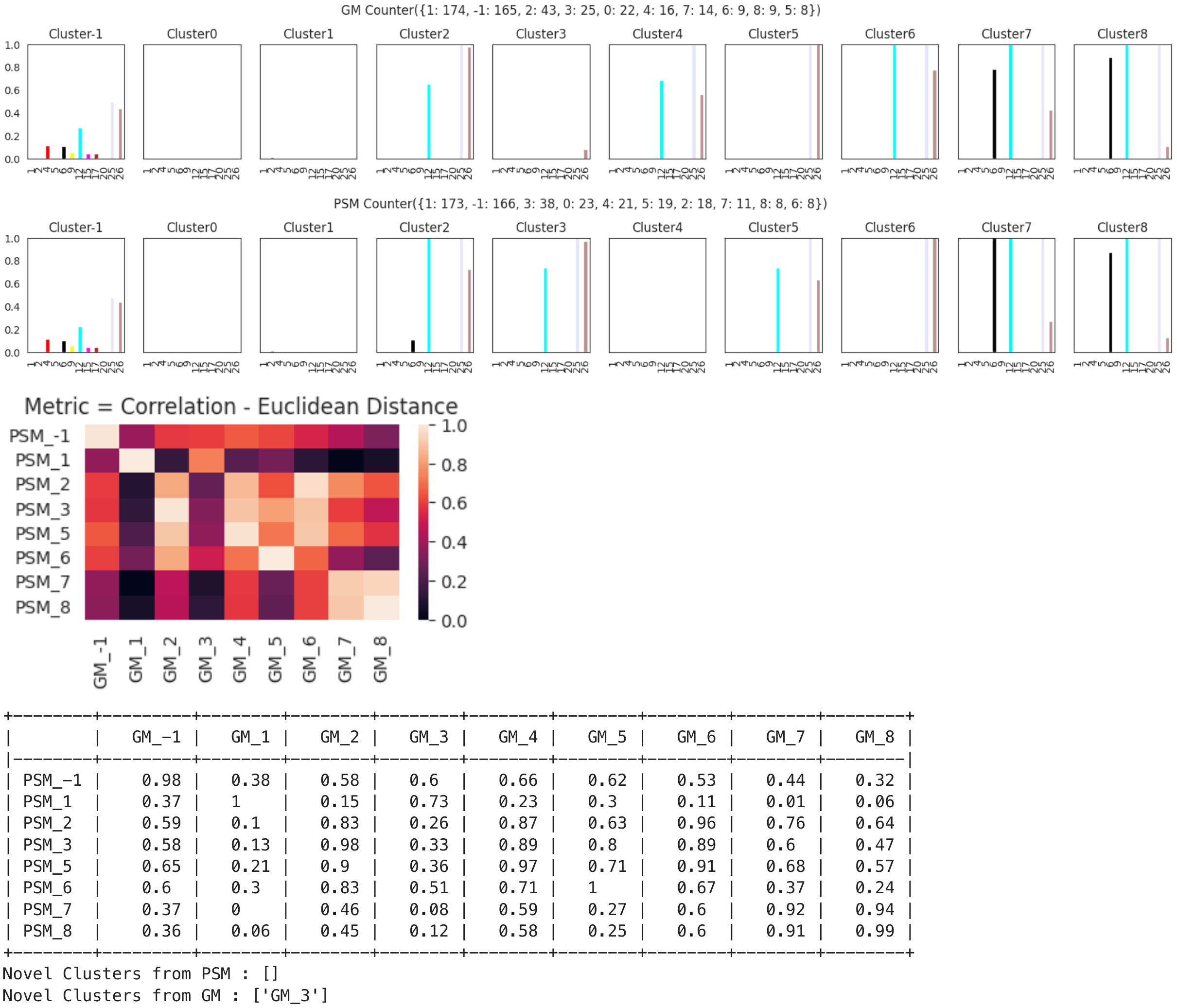}
   \vspace*{-4mm}
   \caption{SN031}
\end{figure*}
\begin{figure*}
\captionsetup{labelformat=empty,labelsep=none}
  \centering
   \includegraphics[width=1\linewidth]{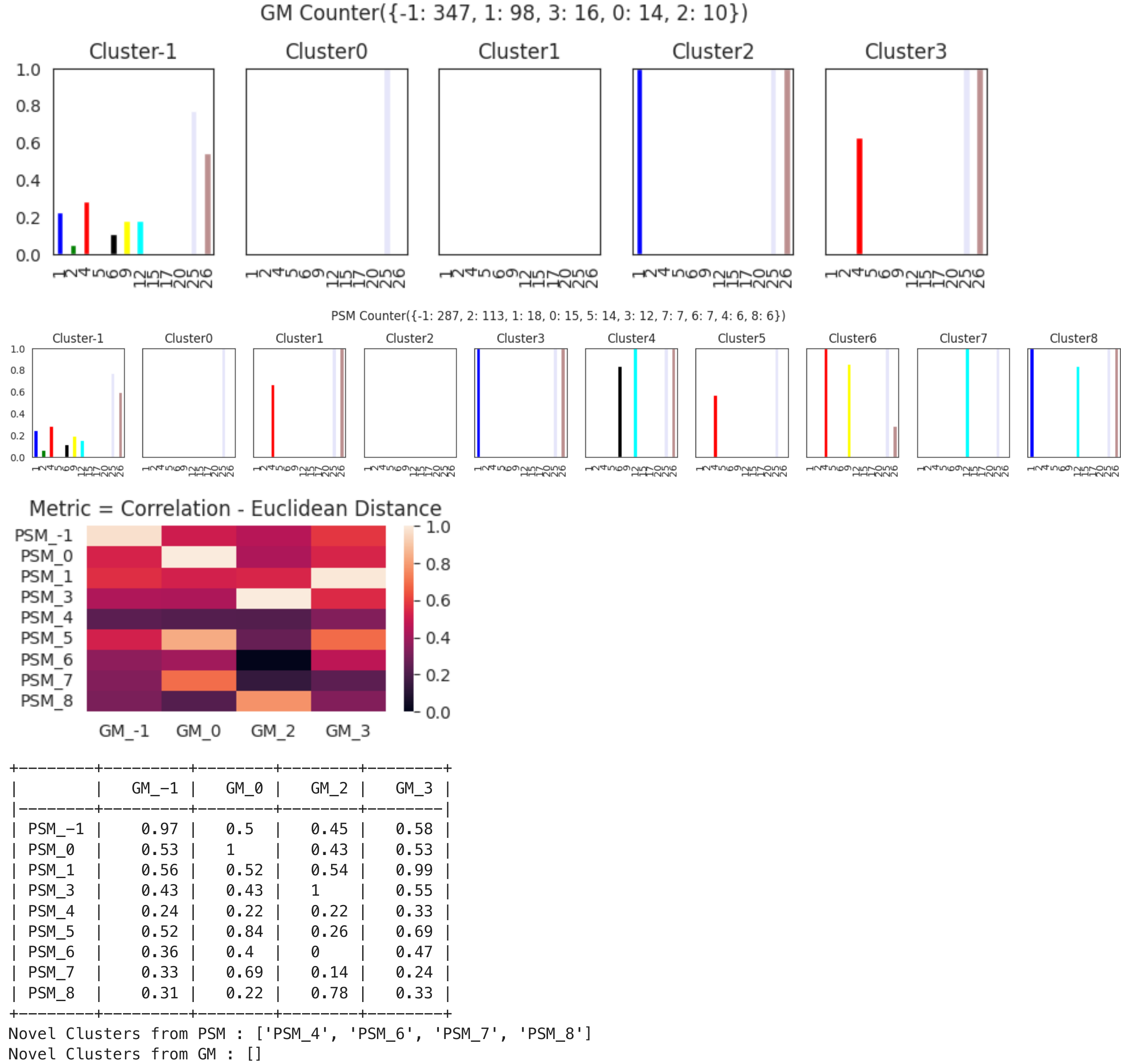}
   \vspace*{-4mm}
   \caption{SN032}
\end{figure*}
\end{document}